\documentclass[letterpaper, 10 pt, conference]{ieeeconf}  % Comment this line out if you need a4paper

\IEEEoverridecommandlockouts                              % This command is only needed if 
\usepackage{graphics} % for pdf, bitmapped graphics files
    \DeclareGraphicsExtensions{.jpg,.pdf,.mps,.png} % for pdflatex
    \graphicspath{{fig/} {./}} %put all figures in these dirs
\usepackage{amsmath} % assumes amsmath package installed
\usepackage{amssymb}  % assumes amsmath package installed
\usepackage{setspace} % used to set te space for long equations
\usepackage{bm} % added by Kentaro UNO 2020/12/08 to use the \bm font
\usepackage[flushmargin,hang]{footmisc} % added by Kentaro UNO 2020/11/05 to make the letter of \thanks to aligned to left
\usepackage{flushend} % added by Kentaro UNO 2022/02/28

\newcommand{\fig}[1]{Fig.~\ref{#1}}

\newcommand{\tab}[1]{Table~\ref{#1}}

\newcommand{\eq}[1]{(\ref{#1})}

\def\epsgaiji#1{\leavevmode\kern-0.025zw\raise-.37zh\hbox{%
  \epsfile{file=#1,width=1.05zw}}\kern-0.025zw}
\newcommand{\MARU}[1]{{\ooalign{\hfil#1\/\hfil\crcr\raise.167ex\hbox{\mathhexbox20D}}}}

\usepackage{array}

\usepackage{siunitx} % to use \SI command to visualize the unit
\usepackage{gensymb} % to use \degree sign
\usepackage{multirow} % to use multirow in tables
\usepackage{caption}
\usepackage{subcaption}
\usepackage{colortbl} % table coloring
\usepackage{xcolor} % table coloring

\usepackage{soul}
\usepackage{framed} % to add frames around comments
\usepackage{pgfplots}
\pgfplotsset{compat=newest}
\usetikzlibrary{plotmarks}
\usetikzlibrary{arrows.meta}
\usepgfplotslibrary{patchplots}
\usepackage{xcolor}

\definecolor{mygreen}{rgb}{0.0,1.0,0}
\definecolor{mylightgreen}{rgb}{0.7,0.9,0.0}
\definecolor{myyellow}{rgb}{1.0,1.0,0.2}
\definecolor{myorange}{rgb}{1.0,0.5,0}
\definecolor{myred}{rgb}{1.0,0,0}

\usepackage{grffile}
\pgfplotsset{plot coordinates/math parser=false}
\newlength\fwidth
\newlength\fheight

\usepackage{cite}

\newcommand{\vecb}[2]{ \bm{#1}_{\mathrm{#2}} }
\newcommand{\vecu}[2]{ \bm{#1}^{\mathrm{#2}} }

\newcommand{\auths}[1]{{#1} {$et$ $al.$}}
\newcommand{\figs}[2]{Figs.~{\ref{#1}-\ref{#2}}}

\title{\LARGE \bf
Space Debris Reliable Capturing \\ by a Dual-Arm Orbital Robot: Detumbling and Caging}

\author{Akiyoshi Uchida$^{1*}$, Kentaro Uno$^{1}$ and Kazuya Yoshida$^{1}$% <-this % stops a space
\thanks{$^{1}$A. Uchida, K. Uno, and K. Yoshida are with the Space Robotics Lab. (SRL) in the Department of Aerospace Engineering, Graduate School of Engineering, Tohoku University, Sendai 980-8579, Japan.}%
\thanks{E-mail: {\tt\small uchida.akiyoshi.s3@dc.tohoku.ac.jp}}
\thanks{$^{*}$\textit{Corresponding author is Akiyoshi Uchida.}}
}

\begin{document}

\maketitle
\thispagestyle{empty}
\pagestyle{empty}

%%%%%%%%%%%%%%%%%%%%%%%%%%%%%%%%%%%%%%%%%%%%%%%%%%%%%%%%%%%%%%%%%%%%%%%%%%%%%%%%

\begin{abstract}
% Mounting a robotic arm on the chaser satellite enables one to capture the space debris and manipulate it for more advanced missions such as refueling or deorbiting. For the chaser sat with multiple robot arms, to simplify the control, a capturing method via \textit{caging} was suggested. \textit{Caging} is an idea to geometrically restrict a target object. However, if the debris is tumbling at high speed, the risk of ejection or satellite destruction is increased with direct caging, so the target should be detumbled before the capturing. To deal with such problems, we suggest a repeated contact-based technique with impedance control to mitigate the target's momentum. In this paper, we analyze the proposed detumbling technique from the aspect of impedance parameters. We investigate the effect of them by conducting parametric analysis and demonstrate the successful detumbling and caging sequence of a microsatellite as a representative of space debris. We show contact force during the detumbling sequence was decreased compared with direct caging. The proposed detumbling and caging sequence was validated by simulation and experiments using a dual-arm air floating robot in a two-dimensional microgravity emulating testbed. 

A chaser satellite equipped with robotic arms can capture space debris and manipulate it for use in more advanced missions such as refueling and deorbiting. To facilitate capturing, a caging-based strategy has been proposed to simplify the control system. \textit{Caging} involves geometrically constraining the motion of the target debris, and is achieved via position control. However, if the target is spinning at a high speed, direct caging may result in unsuccessful constraints or hardware destruction; therefore, the target should be de-tumbled before capture. To address this problem, this study proposes a repeated contact-based method that uses impedance control to mitigate the momentum of the target. In this study, we analyzed the proposed detumbling technique from the perspective of impedance parameters. We investigated their effects through a parametric analysis and demonstrated the successful detumbling and caging sequence of a microsatellite as representative of space debris. The contact forces decreased during the detumbling sequence compared with direct caging. Further, the proposed detumbling and caging sequence was validated through simulations and experiments using a dual-arm air-floating robot in two-dimensional microgravity emulating testbed.

\end{abstract}

%%%%%%%%%%%%%%%%%%%%%%%%%%%%%%%%%%%%%%%%%%%%%%%%%%%%%%%%%%%%%%%%%%%%%%%%%%%%%%%%

\section{INTRODUCTION}
\label{sec:intro}

% ・RSJの序論をベースに，これまでのSRLでの取り組みの内容＋内輪の先行研究だけでなく他の研究機関の取り組みを網羅的にまとめる．

% ・今回のキモはRepeated Impact + Caging + Admittance Controlをすべて盛り込んで実験検証まで行ったことになる

% \todo{Important: to put an appealing fig. 1 at the top right corner to describe the whole picture of this work at a glance.}

% \todo{We should change the order of the intro: background, motivation, and contribution}

\subsection{Background}
\label{subsec:background}

% debris problem 
In recent years, the amount of space debris in Earth's orbit has increased exponentially owing to the increased demand for rocket launches, which has resulted in the deployment of upper stages and an increased number of satellites reaching the end of their operational lives~\cite{EsaDebris2023}. Such debris orbits at high speeds and can cause serious damage if they collide with other satellites or spacecrafts under operation. In February 2009, the U.S. communications satellite Iridium 33 was destroyed when it collided with the Russian communications satellite Cosmos 2251, which was no longer in operation~\cite{Kelso2009}. These accidents increase the amount of space debris, which is a major obstacle to future space development and exploration~\cite{kessler2010}. Thus, the establishment of space debris removal methods, known as active debris removal (ADR), is essential for sustainable space utilization.

% ADR detail
Several institutions are working towards the realization of ADR. For example, the Japan Aerospace Exploration Agency (JAXA) is leading the Commercial Removal of Space Debris Demonstration (CRD2)~\cite{yamamoto2021pave}.
This project has been divided into two phases: Phase I, approaching and inspecting debris, which includes spacecraft pose estimation; and Phase I\hspace{-1.2pt}I, debris capture and de-orbit. As in CRD2 Phase I, spacecraft pose estimation prior to debris capture is essential. \auths{Price} utilized a convolutional neural network (CNN) to estimate the spacecraft pose and demonstrated its efficiency~\cite{andrew2023}. 

\begin{figure}[t]
% \renewcommand{\baselinestretch}{0.6}
% \vspace{-1mm}
  \centering
  \includegraphics[width=\linewidth]{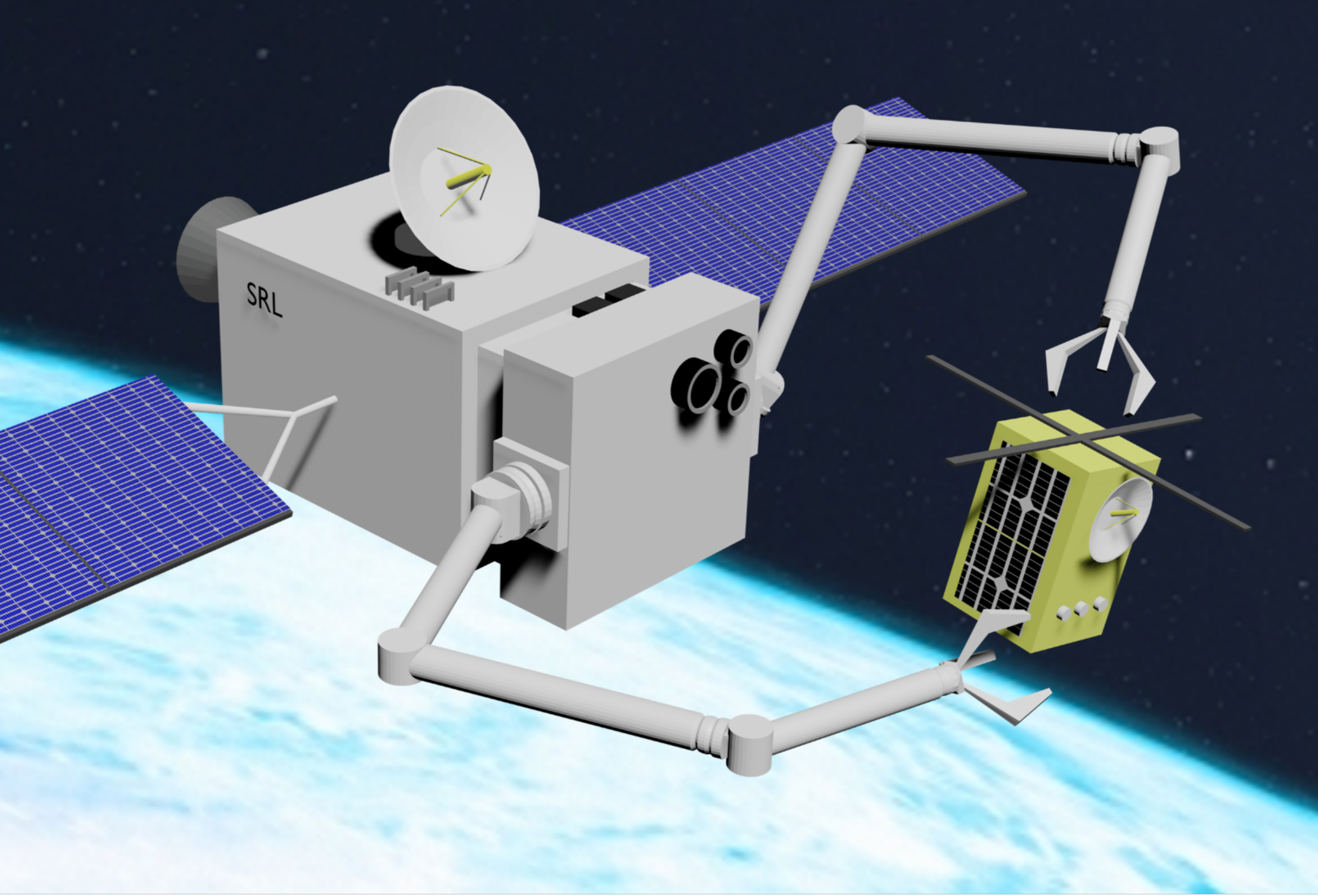}
  \vspace{-5mm}\caption{Conceptual illustration of space debris capture via a dual-arm orbital robot.}\label{fig:scenario}
% \vspace{-5mm}
\end{figure}

% why robotic arm
Several ADR methods have been proposed~\cite{Shan2016}. One method involves capturing debris using manipulators~\cite{Papadopoulos2021}. This method offers the advantage of handling debris with high precision compared to other methods, such as catching debris with a net~\cite{wormnes2013esa} or deorbiting it using an electrodynamic tether~\cite{ISHIGE2004}. Thus, it has high applicability after capture, such as repairing or refueling (see~\fig{fig:scenario}). However, if the contact force when capturing debris is high, the risk of serious damage to the servicing satellite increases, along with the possibility of debris ejection. Therefore, force-aware control is necessary for the reliable capture of space debris.

In general, impedance control~\cite{Hogan1985} efficiently reduces the contact force between the target and end effectors. To reduce the contact force, \auths{Nakanishi} applied impedance control to a satellite model with a single manipulator and simulated the capture sequence of a target approaching translationally~\cite{Nakanishi2006}. This study demonstrated the effectiveness of impedance control in a microgravity environment. In addition, \auths{Abiko} validated the performance of impedance control in grasping a target with parameter uncertainty through simulation~\cite{Abiko2006}. Furthermore, \auths{Dongming} defined a dynamic model for a service satellite with multiple manipulators wherein impedance control was applied during the capture of the target satellite~\cite{Dongming2020}. 
%Furthermore, as an example of application to a mission, they demonstrated the effectiveness of the method by simulating the direct grasping of a tumbling target by a dual-arm robot testbed.

Another approach for the reaction-aware capturing of tumbling debris by two manipulators was proposed by \auths{Nagaoka}. This involved the capturing of a rotating target by repeatedly touching it with dual manipulators~\cite{Nagaoka2018}. The effectiveness of the method was experimentally verified by capturing a cylindrical target using a robot model with spherical end effectors. In this method, a rotating target was detumbled by repeated impacts and captured without force control. The use of two arms improved the safety of the sequence by allowing immediate contact with debris that was sent flying in the opposite direction of the arm owing to contact. They confirmed that repeated impacts reduced the target motion gradually. 
%A dual-arm air-floating robot, which has been developed in Space Robotics Laboratory was used in this research.

The active grasping of debris requires continuous force control; thus, \textit{caging} was proposed to simplify the problem. Caging geometrically restricts an object; thus, no complex or precise grasping force control is required. \auths{Matsushita} utilized a U-shaped end effector to a dual-arm robot that enabled caging~\cite{Matsushita2020}, based on the theory of object closure~\cite{Dong2002}. They demonstrated that the end effector was efficient in caging relatively slow spinning targets of circular and square shapes with simple position control.

% in lab research 
% In this laboratory, we have focused on space debris removal methods using manipulators and have conducted research using a dual-arm robot testbed~\cite{fig:dar}. Nagaoka et. al. proposed to detumble and capture a rotating target using Repeated Impact-Based Capture. They verified the effectiveness of the method experimentally by capturing a cylindrical target using a robot model with spherical end effectors~\cite{RIC}. Matsushita et al. applied a U-shaped end effector that enables caging, based on the theory of object closure, to a dual-arm robot testbed and demonstrated the method is efficient in capturing the floating target with relatively simple control. Caging is an idea to geometrically restrict an object by using robotic manipulators, thus no complicated and precise grasping force control was required.

% In this research, we combined impedance control and the Repeated Impact-Based Capture method to propose a more reliable sequence to detumble debris whose shape is rectangular. Furthermore, we verified the comprehensive sequence to capture the debris using caging method after the detumbling. The result was validated by both simulation and experiment.

\subsection{Motivation}
\label{subsec:motivation}
The aforementioned approaches have been validated independently; however, to reliably detumble and capture the target with complex and rapid motion, the techniques of multi-contact, caging, and force control must be optimally integrated for the following reasons. Single contact-based caging is excessively risky in cases where 1) the target tumbling motion is fast and complicated and 2) the target mass is unknown. The complexity of the target motion necessitates high-precision control. Furthermore, the uncertainty of the inertial properties of the target may cause a rapid transfer of momentum to the manipulator, which could overcome the control capability of the chaser satellite or create new debris in a worst-case scenario. Therefore, a force-controlled multiple-contact-based detumbling phase~\cite{WANG2020} is essential for reducing the rotation speed and stabilizing the orientation for caging. Moreover, the preliminary contacts facilitate the estimation of the target inertial properties to refine the impedance parameters for eventual capture.

% However, if the target is tumbling at a high speed, the possibility of ejecting or satellite destruction is increased if the manipulator hits debris simply. Therefore, the impact force between the end effector and debris needs to be decreased. Furthermore, the method proposed by \auths{Nagaoka} did not deal with square-shaped targets, which are the shape of many artificial satellites, so expanding their research in terms of target shape is also required.

% \begin{figure}[t]
%     \begin{tabular}{cc}
%       \begin{minipage}[t]{0.9\hsize}
%          \centering
%          \includegraphics[width=0.9\linewidth]{fig/DAR_o.pdf}
%          \vspace{-2mm}
%          \subcaption{~\cite{xxxx}}
%          \label{fig:dar_o}
%      \end{minipage} \vspace{2mm} \\
%       \begin{minipage}[t]{0.9\hsize}
%          \centering
%          \includegraphics[width=0.9\linewidth]{fig/DAR_u.pdf}
%          \vspace{-2mm}
%          \subcaption{~\cite{xxxx}}
%          \label{fig:dar_u}
%      \end{minipage}
%     \end{tabular}
%     \vspace{-2mm}
%     \caption{}
%     \label{fig:dar}
%     \vspace{-5mm}
% \end{figure}

\subsection{Contributions}
\label{subsec:contributions}
In this study, we proposed a new reliable sequence for detumbling and capturing rotating space debris by combining the aforementioned technical elements. The primary contributions of this study are as follows.

\begin{itemize}
    \item We expanded the repeated impact-based capture~\cite{Nagaoka2018} for detumbling quadrangular prism targets and rendered it more reliable by using impedance control.
    \item A parametric simulation was conducted to investigate the effect of impedance parameters on the capturing result.
    \item We demonstrated a comprehensive sequence to capture debris using a dual-arm orbital robot, by capturing the target using the caging method following the preliminary detumbling phase.
\end{itemize}

\section{DUAL-ARM ORBITAL ROBOT}
\label{sec:dual_arm_robot}
\subsection{Robot model}
\begin{figure}[t]
% \renewcommand{\baselinestretch}{0.8}
% \vspace{0mm}
  \centering
  \includegraphics[width=.75\linewidth]{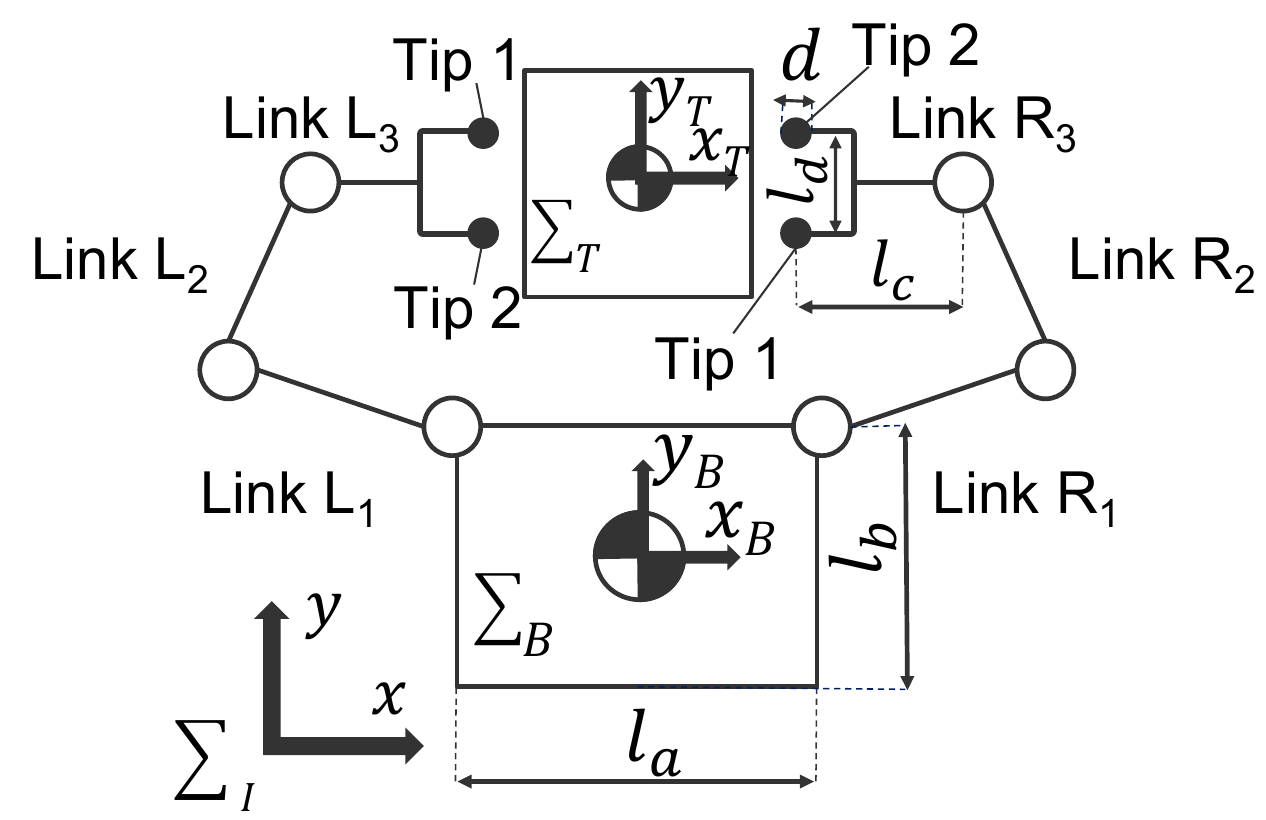}
  \vspace{0mm}\caption{Dual-arm orbital robot and target models. }\label{fig:sim_model}
% \vspace{-5mm}
\end{figure}

\fig{fig:sim_model} shows the robot and target model used in this research. As shown in this figure, the chaser robot comprised a base and six links, with three links on each of the two arms. Further, each arm had a U-shaped end effector with two end-tip spheres. These two arms allowed the robot to repeatedly contact and grasp the target while reducing the risk of pushing the target away. In addition, the motion of the robot was restricted to a plane by six parallel joints, although the following equations are discussed in three-dimensional space. 

\subsection{Basic equations}
\subsubsection{Equation of motion}

The space robot used in this study had two articulated arms~\cite{Yoshida1991}. Let $k=L$ and $k=R$ be the symbols for the left and right arms, respectively. Furthermore, the number of joints in arm $k$ is defined as ${}^kn$. The equation of motion for a dual-arm space robot is expressed as \eq{eq:dar_dyn}. The variables defined here were assumed to be defined in the inertial coordinate system $\Sigma_I$ unless otherwise noted. 

\begin{spacing}{0.5}
\begin{align*} % definition
    \vecb{H}{b} &\in \vecu{R}{} {}^{6 \times 6}& &\text{: Inertia matrix of the entire system}\\
    {}^{k} \vecb{H}{m} &\in \vecu{R}{} {}^{{}^kn \times {}^kn}& &\text{: Inertia matrix of the arm $k$}\\
    {}^{k} \vecb{H}{bm} &\in \vecu{R}{} {}^{6 \times {}^kn}& &\text{: Interference matrix of the base}\\
    & & &\text{and the arm $k$}\\
    \vecb{x}{b} &\in \vecu{R}{6}& &\text{: Pose of the COG of the base}\\
    {}^{k} \vecb{\phi}{} &\in \vecu{R}{} {}^{{}^kn}& &\text{: Joint angle of the arm $k$}\\
    \vecb{c}{b} &\in \vecu{R}{6}& &\text{: Non-linear velocity term of the base}\\
    {}^{k} \vecb{c}{m} &\in \vecu{R}{} {}^{{}^kn}& &\text{: Non-linear velocity term of the arm $k$}\\
    \vecb{F}{b} &\in \vecu{R}{6}& &\text{: External force applied to the base}\\
    {}^{k} \vecb{\tau}{} &\in \vecu{R}{} {}^{{}^kn}& &\text{: Joint torque of the arm $k$}\\
    {}^{k} \vecb{J}{b} &\in \vecu{R}{} {}^{6 \times 6}& &\text{: Jacobian of the base}\\
    {}^{k} \vecb{J}{m} &\in \vecu{R}{} {}^{6 \times  {}^kn}& &\text{: Jacobian of the arm $k$}\\
    {}^{k} \vecb{F}{h} &\in \vecu{R}{6}& &\text{: External force applied to the}\\
    & & &\text{end effector of the arm $k$}\\
\end{align*}
\end{spacing}

\begin{spacing}{0.5}
\begin{equation}
 \begin{split}
    \begin{bmatrix}
        \vecb{H}{b} & {}^{L} \vecb{H}{bm} & {}^{R} \vecb{H}{bm} \\ {}^{L} \vecb{H}{bm} \vecu{}{T} & {}^{L} \vecb{H}{m} &  \vecb{0}{} {}_{ {}^Ln \times  {}^Rn}\\
        {}^{R} \vecb{H}{bm} \vecu{}{T} & \vecu{0}{T} {}_{ {}^Ln \times  {}^Rn} & {}^{R} \vecb{H}{m}
    \end{bmatrix}
    \begin{bmatrix}
        {\vecb{\ddot{x}}{b}}\\
        {{}^{L} \vecb{\ddot{\phi}}{}}\\
        {{}^{R} \vecb{\ddot{\phi}}{}}
    \end{bmatrix}
    +
    \begin{bmatrix}
        \vecb{c}{b}\\
        {}^{L} \vecb{c}{m}\\
        {}^{R} \vecb{c}{m}
    \end{bmatrix}\\
    = 
    \begin{bmatrix}
        \vecb{F}{b}\\
        {}^{L} \vecb{\tau}{}\\
        {}^{R} \vecb{\tau}{}
    \end{bmatrix}
    +
    \begin{bmatrix}
        {}^{L} \vecb{J}{b} \vecu{}{T} & {}^{R} \vecb{J}{b} \vecu{}{T}\\
        {}^{L} \vecb{J}{m} \vecu{}{T}  & \vecu{0}{T} {}_{6 \times  {}^Ln}\\
        \vecu{0}{T} {}_{6 \times  {}^Rn} & {}^{R} \vecb{J}{m} \vecu{}{T} 
    \end{bmatrix}
    \begin{bmatrix}
        {}^{L} \vecb{F}{h}\\
        {}^{R} \vecb{F}{h}
    \end{bmatrix}
 \end{split}
 \label{eq:dar_dyn}
 \vspace{-3mm}
\end{equation}
\end{spacing}

\subsubsection{Generalized Jacobian}
To control the end effector velocity of the robot under a microgravity environment, generalized Jacobian, which was derived by \auths{Umetani}~\cite{Umetani1989}, was used. The relationship between the joint and end effector velocities is shown in \eq{eq:g_jacob}. By solving this equation for $\dot{\vecb{\phi}{}}$, the target joint velocity can be calculated using the desired end effector velocity.

\begin{spacing}{0.5}
\begin{align*}
    {}^{k}\vecb{x}{h} &\in \vecu{R}{6}& &\text{: Pose of the end effector of the arm $k$}\\
    \vecu{J}{*} &\in \vecu{R}{} {}^{6 \times ( {}^Ln +  {}^Rn)}& &\text{: Generalized jacobian}\\
    \vecb{P}{} &\in \vecu{R}{6}& &\text{: Momentum of the robot system}
\end{align*}
\end{spacing}

\begin{equation}
 \begin{split} 
    \begin{bmatrix}
        &{}^{L}\vecb{\dot{x}}{h} &- &{}^{L} \vecb{J}{b} \vecb{H}{b} {}^{-1} \vecb{P}{}\\
        &{}^{R}\vecb{\dot{x}}{h} &- &{}^{R} \vecb{J}{b} \vecb{H}{b} {}^{-1} \vecb{P}{}
    \end{bmatrix}
    = \vecu{J}{*} 
    \begin{bmatrix}
        {}^{L} \vecb{\dot{\phi}}{}\\
        {}^{R} \vecb{\dot{\phi}}{}
    \end{bmatrix}
 \end{split}
 \label{eq:g_jacob}
\end{equation}
\begin{equation*}
    \vecu{J}{*} = 
    \begin{bmatrix}
        {}^{L} \vecb{J}{m} - {}^{L} \vecb{J}{b} \vecb{H}{b} {}^{-1} {{}^{L} \vecb{H}{bm}} &
        - {}^{L} \vecb{J}{b} \vecb{H}{b} {}^{-1} {{}^{R}\vecb{H}{bm}}\\
        - {}^{R} \vecb{J}{b}\vecb{H}{b}{}^{-1} {{}^{L}\vecb{H}{bm}} &
        {}^{R} \vecb{J}{m} - {}^{R} \vecb{J}{b}\vecb{H}{b} {}^{-1} {{}^{R}\vecb{H}{bm}}
    \end{bmatrix}
\end{equation*}
\subsubsection{Impedance control}
The fundamental equation of impedance control is expressed as \eq{eq:imp}, where $\vecb{M}{im} \in \vecu{R}{6 \times 6}$, ${\vecb{D}{im} \in \vecu{R}{6 \times 6}}$, and $\vecb{K}{im} \in \vecu{R}{6 \times 6}$ are the virtual mass, damping coefficient, and elastic coefficient that determine the impedance characteristics, respectively. In addition, $\Delta^{k}\vecb{x}{h}$ represents the displacement of arm $k$ from the equilibrium point. In this study, the equilibrium position is the position of the end effector at the point of first contact with the target. By solving this equation in terms of velocity, the end effector velocity of the free-flying robot was controlled via a generalized Jacobian using~\eq{eq:g_jacob}. Therefore, the method used in this study is a virtual internal model-following control~\cite{Kosuge1987}, also referred to as \textit{admittance control}. 
\begin{equation}
    \vecb{M}{im} \Delta^{k} \vecb{\ddot{x}}{h} + \vecb{D}{im} \Delta^{k} \vecb{\dot{x}}{h} + \vecb{K}{im} \Delta^{k} \vecb{x}{h} 
    = {}^{k}\vecb{F}{h}
    \label{eq:imp}
\end{equation}

Admittance control feeds back the force sensor values, which delays computation. Thus, if the contact time is excessively short, control cannot be performed in time and the impact cannot be sufficiently reduced. Therefore, for effective impedance control, a compliance mechanism or flexible end effector must be used to increase the contact time~\cite{Uyama2012}. The robot used in this study handled this problem by installing a soft material at the tip of the end effector.

\section{DEBRIS CAPTURING SEQUENCE}
\label{sec:results}
\subsection{Problem formulation}
The debris-capturing mission was modeled as shown in \fig{fig:cap_seq}. The following conditions were assumed for the model:
\begin{itemize}
    \item All motion was limited to a two-dimensional plane.
    \item The robot began the sequence under the condition that the approach to the target had been completed.
    \item The target shape was assumed to be a quadrangular prism as modeling cubic satellites.
    \item The initial momentum of the robot was zero.
    \item The target rotated without translation at the initial state of the sequence.
    \item Gravitational acceleration was assumed to be 0
    \item Contact was only allowed between the end effectors' sphere tips and the target.
    \item We assumed that the robot knows the shape and motion state of the target.
\end{itemize}
\subsection{Debris detumbling and capture strategy}
\begin{figure}[t]
% \renewcommand{\baselinestretch}{0.8}
% \vspace{-1mm}
  \centering
  \includegraphics[width=.95\linewidth]{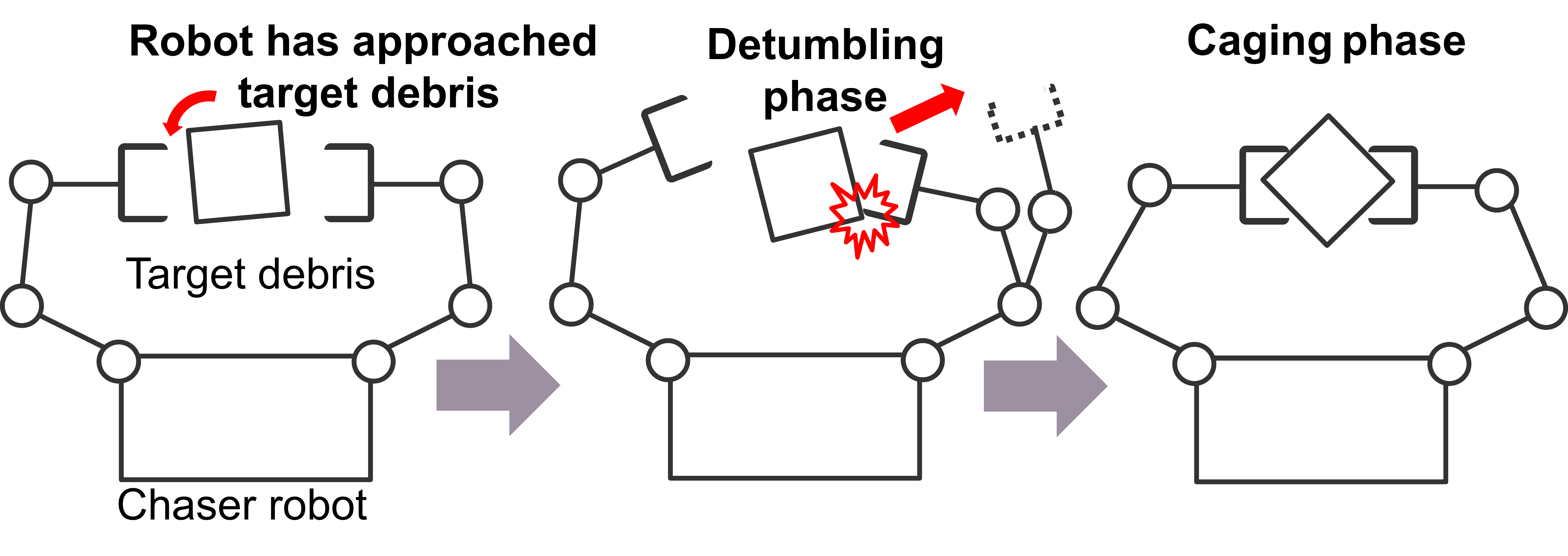}
  \vspace{0mm}\caption{Detumbling and capturing sequence.}\label{fig:cap_seq}
% \vspace{-5mm}
\end{figure}

The space debris-capturing flow is shown in \fig{fig:cap_seq}. Once the robot began the sequence under the condition that the approaching phase was completed, the target motion was damped by repeated one-arm contacts with impedance control until the angular velocity of the target became smaller than a predetermined threshold. During the detumbling phase, the end effector followed the trajectory until it hit the target. If the force applied to the end effector was beyond the threshold, implying that the target hit, the robot switched its control mode from the trajectory-following to the impedance-based reduction mode until an empirically predetermined threshold of time was passed for the manipulator to adequately absorb the shock. Time, rather than force, was selected as the threshold because the end effector stopped excessively close to the target after impact, and the risk of unwanted contact was greater if the force threshold was employed. When the target angular velocity was below the threshold value, the target was captured using caging. After caging, the angular velocities of all joints were set to zero, and the arm was fixed with respect to the floating robot base.

\begin{figure}[t]
    \begin{tabular}{cc}
     \begin{minipage}[t]{0.45\hsize}
         \centering
         \includegraphics[width=1.1\linewidth]{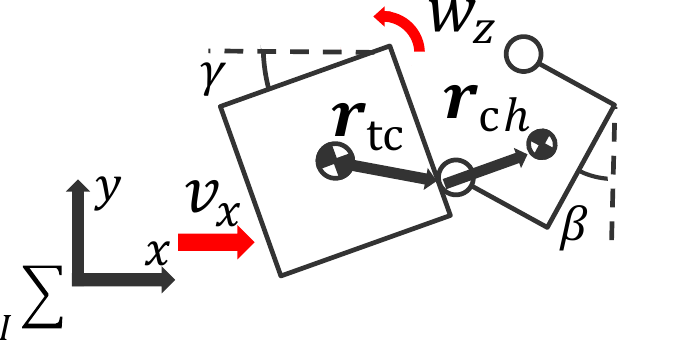}
         \subcaption{$v_x>0, \omega_z>0$}
         \label{fig:cont_pat_a}
     \end{minipage}&  
      \begin{minipage}[t]{0.45\hsize}
         \centering
         \includegraphics[width=0.8\linewidth]{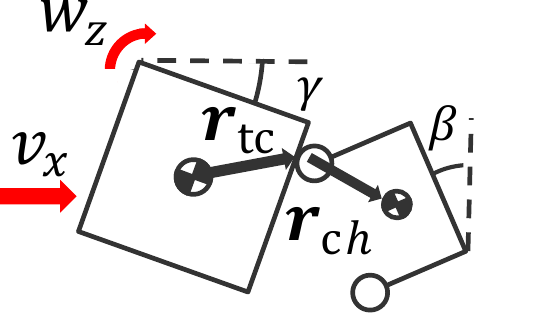}
         \subcaption{$v_x>0, \omega_z<0$}
         \label{fig:cont_pat_b}
     \end{minipage} \vspace{3mm} \\ 
      \begin{minipage}[t]{0.45\hsize}
         \centering
         \includegraphics[width=0.8\linewidth]{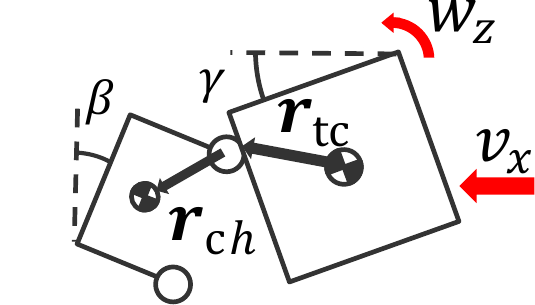}
         \subcaption{$v_x<0, \omega_z>0$}
         \label{fig:cont_pat_c}
     \end{minipage}&
      \begin{minipage}[t]{0.45\hsize}
         \centering
         \includegraphics[width=0.8\linewidth]{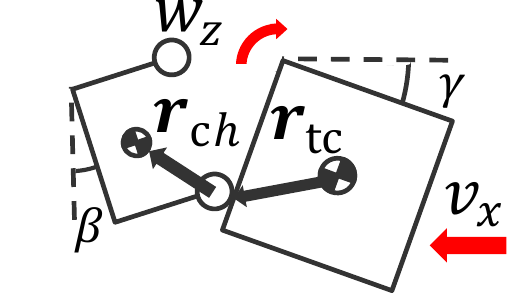}
         \subcaption{$v_x<0, \omega_z<0$}
         \label{fig:cont_pat_d}
     \end{minipage}
    \end{tabular}
    \vspace{0mm}\caption{Contact patterns for target detumbling. }
    \label{fig:contact_mode} 
    % \vspace{-1mm}
\end{figure}

\subsection{End effector trajectory planning}
\label{subsec:traject_plan}
\subsubsection{Goal point}
\label{subsubsec:goal_point}
To achieve the optimal goal point (i.e., hitting position) of the end effector to attenuate the angular and linear momentum of the target, four contact patterns were defined according to the target motion state, as shown in \fig{fig:contact_mode}. The following variables were newly defined, and the formula for calculating the desired end effector position is shown below, where $\vecb{R}{}(\theta)$ represents the rotation matrix of the rotation angle $\theta$ and $\text{sign}(x)$ is the sign of $x$. In addition, the desired angle of the end effector was determined by $\beta$ in \fig{fig:contact_mode}, which was set as the control parameter. Here, the upper-left subscripts had the following meanings: i : value of the variable at the time of trajectory design. g: desired value of variable at the end of the trajectory or goal value. d: desired value of variable during trajectory. e: error value of variable during trajectory.

\begin{spacing}{0.5}
\begin{align*}
    t &\in \vecu{R}{1}& &\text{: Elapsed time during the trajectory}\\
    T &\in \vecu{R}{1}& &\text{: Duration of the trajectory}\\
    \vecb{p}{h} &\in \vecu{R}{2}& &\text{: Position of the end effector}\\
    \vecb{v}{h} &\in \vecu{R}{2}& &\text{: Velocity of the end effector}\\
    \theta_{\mathrm{h}} &\in \vecu{R}{1}& &\text{: Angle of the end effector}\\
    \omega_{\mathrm{h}} &\in \vecu{R}{1}& &\text{: Angular velocity of the end effector}\\
    \vecb{p}{t} &\in \vecu{R}{2}& &\text{: Position of the target}\\
    \vecb{v}{t} &\in \vecu{R}{2}& &\text{: Velocity of the target}\\
    u_\mathrm{t} &\in \vecu{R}{1}& &\text{: $x$ component of the target velocity}\\
    \theta_{\mathrm{t}} &\in \vecu{R}{1}& &\text{: Angle of the target}\\
    \omega_\mathrm{t} &\in \vecu{R}{1}& &\text{: Angular velocity of the target}\\
    {}_{t} \vecb{r}{tc}&\in \vecu{R}{2}& &\text{: Position of the contact point expressed in}\\
    & & &\text{target frame}\\
    \vecb{r}{ch} &\in \vecu{R}{2}& &\text{: Relative position of the end effector from}\\
    & & &\text{contact point}\\
    l_{\mathrm{t}} &\in \vecu{R}{1}& &\text{: Length of the target side}
\end{align*}
\begin{equation}
    \vecu{}{g} \vecb{p}{h} = \vecu{}{i} \vecb{p}{t} + \vecu{}{i} \vecb{v}{t}T + \vecb{R}{}(\vecu{}{i} \theta_{\mathrm{t}} + \vecu{}{i} \omega_\mathrm{t} T) {}_{t} \vecb{r}{tc} + \vecb{r}{ch}
    \label{eq:des_pos}
    \vspace{2mm}
\end{equation}
\end{spacing}

Here, ${}_{t} \vecb{r}{tc}$ is dependent on the state of motion of the target, as shown in \fig{fig:contact_mode}. By defining the contact position as \eq{eq:cont_mode}, the contact force can be applied in a direction that simultaneously dampens the rotational and translational motions of the target. Note that $\alpha \in [0, 1]$ is a dimensionless parameter that indicates the contact position on the target edge, where zero indicates the center of the target edge and one indicates the target apex. 

\begin{spacing}{1.0}
\begin{equation}
    {}_{t} \vecb{r}{tc} 
    =
    \begin{bmatrix}
       \frac{l_{\mathrm{t}}}{2} \mathrm{sign}(\vecu{}{i} u_\mathrm{t})\\
        -\frac{l_{\mathrm{t}} \alpha}{2} \mathrm{sign}(\vecu{}{i} u_\mathrm{t} \vecu{}{i} \omega_\mathrm{t}) 
    \end{bmatrix}
    \label{eq:cont_mode}
    \vspace{2mm}
\end{equation}
\end{spacing}

\subsubsection{Path generation}
% During the sequence, the robot is controlled by the velocity of the end effector. Thus, it is required to calculate the velocity to achieve the desired goal point of the end effector. 
After calculating the goal points of the end effector according to the strategy in \ref{subsubsec:goal_point}, a path was generated by connecting the goal position and the current end effector position with straight lines between them to achieve position control. ~$s$ is a dimensionless number representing the trajectory position.

\begin{equation}
    \begin{bmatrix}
       \vecu{}{d} \vecb{p}{h}(s) \\
       \vecu{}{d} \theta_h(s)
    \end{bmatrix}
    = 
    \begin{bmatrix}
        \vecu{}{i} \vecb{p}{h} \\
        \vecu{}{i} \theta_h 
    \end{bmatrix}
     + s 
     \begin{bmatrix}
         \vecu{}{g} \vecb{p}{h} - \vecu{}{i} \vecb{p}{h} \\
         \vecu{}{g} \theta_h - \vecu{}{i} \theta_h
     \end{bmatrix},~~~~
     s \in [0, 1]
\end{equation}

\subsubsection{Time scaling}
To calculate the velocity of the end effector, the path must be scaled over time. In this study, 3rd-order polynomial time scaling
%, which is shown in \fig{time_scaling},
 was used (equation defined as follows). Using this time scaling, the velocity of the end effector at the starting and ending points was designed to be zero to avoid discontinuous control.
% \vspace{-2mm}
\begin{equation}
    s(t) = \frac{3}{T^2}t^2 - \frac{2}{T^3}t^3
\end{equation}
% \vspace{-2mm}

% \begin{figure}[t]
%   \centering
%   \includegraphics[width=0.8\linewidth]{fig/time_scaling.pdf}
%   \vspace{-2mm}\caption{Plots of $s(t)$, $\dot s(t)$, and $\ddot s(t)$ for a third-order polynomial time scaling.}\label{time_scaling}
% \end{figure}

\subsection{Trajectory following control}
Both feedforward and feedback controls were performed to follow the trajectory planned in \ref{subsec:traject_plan}. In an environment wherein the robot can be controlled ideally or very precisely, as in the simulation, trajectory control without feedback works well; however, this is not always true in the actual environment. Therefore, the following equation was used in this study. Here, ${\vecb{K}{p}\in\vecu{R}{3}}$, $\vecb{K}{i}\in\vecu{R}{3}$ and $\vecb{K}{d}\in\vecu{R}{3}$ represent feedback gains. Further, the upper left subscripts have the following meanings: i: value of the variable at the time of designing the trajectory. g: Desired value of the variable at the end of the trajectory or goal value. d: desired value of variable during trajectory. e: error value of the variable during the trajectory.

% \vspace{-5mm}
\begin{align} 
     &\begin{bmatrix}
       \vecu{}{d} \vecb{v}{h}(t) \\
       \vecu{}{d} \omega_h(t)
    \end{bmatrix}
    = \dot s(t) 
     \begin{bmatrix}
         \vecu{}{g} \vecb{p}{h} - \vecu{}{i} \vecb{p}{h} \\
         \vecu{}{g} \theta_h - \vecu{}{i} \theta_h
     \end{bmatrix}& \notag\\
     &+
     \vecb{K}{p}
     \begin{bmatrix}
       \vecu{}{e} \vecb{p}{h} \\
       \vecu{}{e} \theta_h
     \end{bmatrix} 
     +
     \vecb{K}{i} \int 
     \begin{bmatrix}
       \vecu{}{e} \vecb{p}{h} \\
       \vecu{}{e} \theta_h
     \end{bmatrix} dt 
     +
     \vecb{K}{d} \frac{d}{dt}
     \begin{bmatrix}
       \vecu{}{e} \vecb{p}{h} \\
       \vecu{}{e} \theta_h
     \end{bmatrix}
\end{align}

\begin{figure*}[ht]
\centering
\vspace{-0mm}
    % \begin{tabular}{cccc}
    %  \begin{minipage}[t]{0.18\hsize} \centering
    %      \includegraphics[width=30mm]{fig/animation/s000.png}
    %      \vspace{-1mm} \subcaption*{\hspace{4mm}Time = 0.0 [s]}
    %  \end{minipage} &
    %  \begin{minipage}[t]{0.18\hsize} \centering
    %      \includegraphics[width=30mm]{fig/animation/s019c1.png}
    %      \vspace{-1mm} \subcaption*{\hspace{4mm}Time = 1.9 [s]}
    %  \end{minipage} &
    %  \begin{minipage}[t]{0.18\hsize} \centering
    %      \includegraphics[width=30mm]{fig/animation/s020.png}
    %      \vspace{-1mm} \subcaption*{\hspace{4mm}Time = 2.0 [s]}
    %  \end{minipage} &
    %  \begin{minipage}[t]{0.18\hsize} \centering
    %      \includegraphics[width=30mm]{fig/animation/s049c2.png}
    %      \vspace{-1mm} \subcaption*{\hspace{4mm}Time = 4.9 [s]}
    %  \end{minipage} \\
    %  \begin{minipage}[t]{0.18\hsize} \centering
    %      \includegraphics[width=30mm]{fig/animation/s051.png}
    %      \vspace{-1mm} \subcaption*{\hspace{4mm}Time = 5.1 [s]}
    %  \end{minipage} \vspace{3mm} &
    %  \begin{minipage}[t]{0.18\hsize} \centering
    %      \includegraphics[width=30mm]{fig/animation/s115c4.png}
    %      \vspace{-1mm} \subcaption*{\hspace{4mm}Time = 11.5 [s]}
    %  \end{minipage} &
    %  \begin{minipage}[t]{0.18\hsize} \centering
    %      \includegraphics[width=30mm]{fig/animation/s132.png}
    %      \vspace{-1mm} \subcaption*{\hspace{4mm}Time = 13.2 [s]}
    %  \end{minipage} &
    %  \begin{minipage}[t]{0.18\hsize} \centering
    %      \includegraphics[width=30mm]{fig/animation/s137.png}
    %      \vspace{-1mm} \subcaption*{\hspace{4mm}Time = 13.7 [s]}
    %  \end{minipage}
    % \end{tabular}
    
    \includegraphics[width=.9\linewidth]{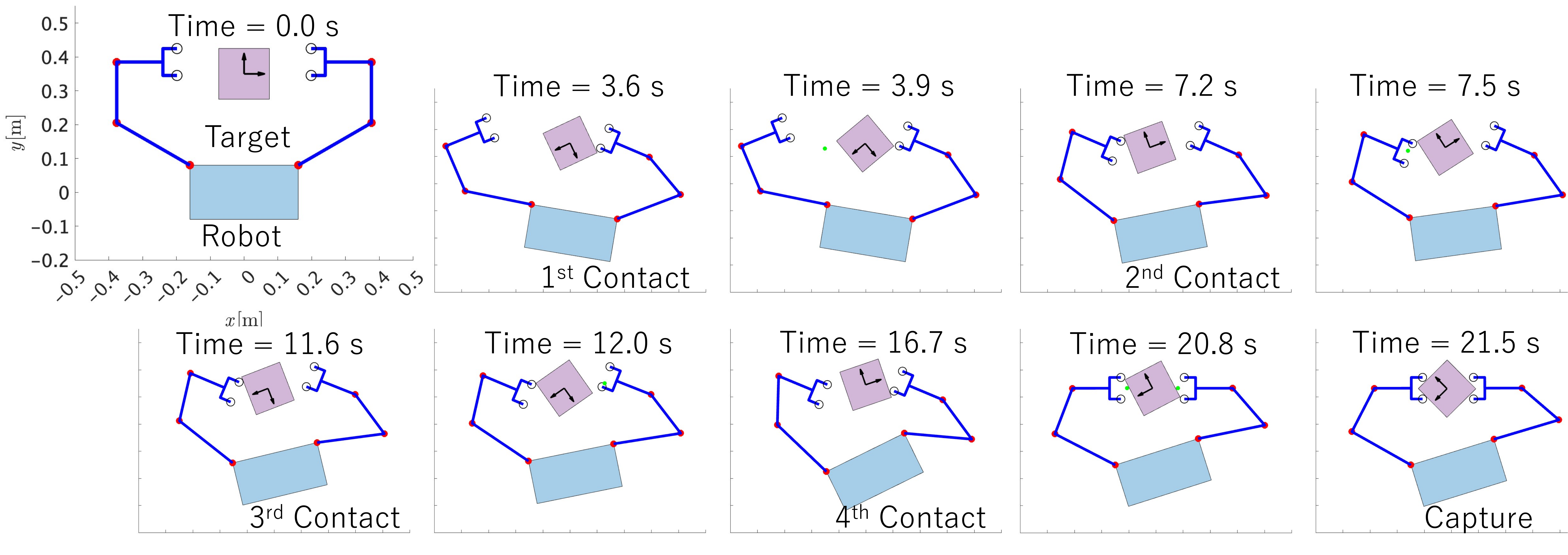}
    % \vspace{-2mm}
    \caption{Snapshots of the space debris capturing simulation.}
    \label{fig:sim_snapshots}
    \vspace{-4mm}
\end{figure*}

\begin{figure}[t]
% \vspace{2mm}
    \begin{tabular}{cc}
     \begin{minipage}[t]{0.45\hsize}
         \centering
         \includegraphics[width=\linewidth]{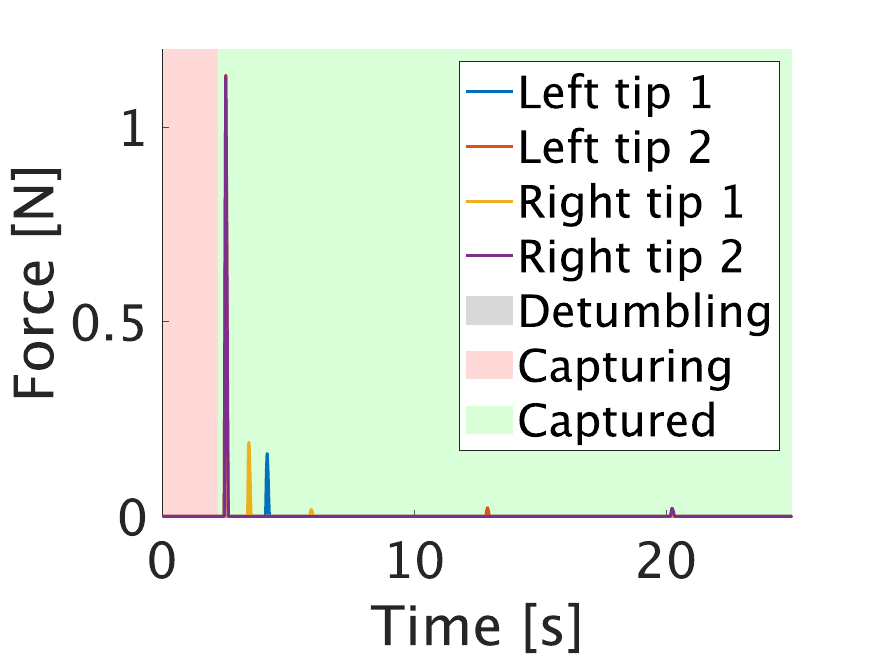}
         \subcaption{Direct caging.}
         \label{fig:sim_dir_f}
     \end{minipage}&  
      \begin{minipage}[t]{0.45\hsize}
         \centering
         \includegraphics[width=\linewidth]{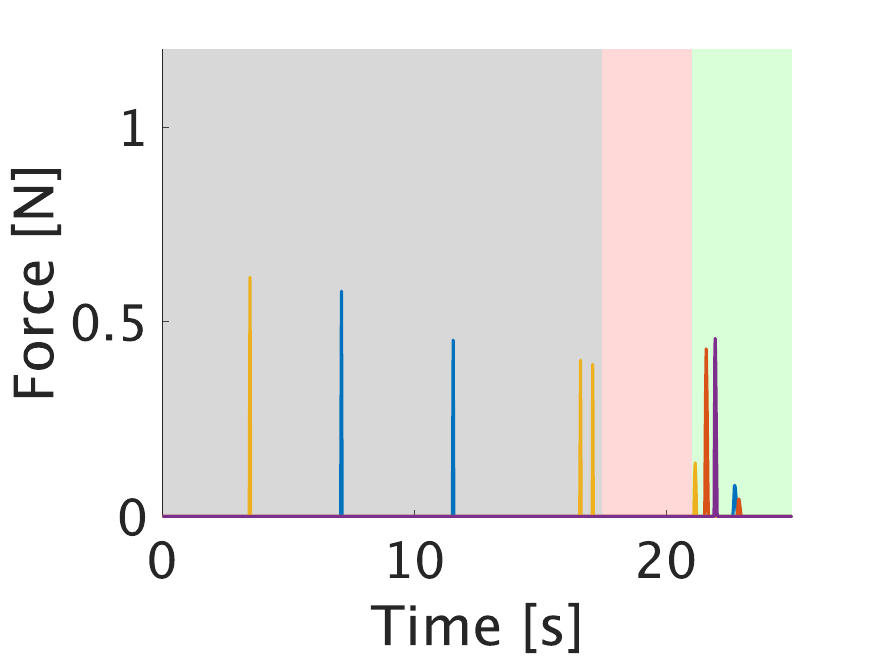}
         \subcaption{Detumbling and caging.}
         \label{fig:sim_mul_f}
     \end{minipage}\\
    \end{tabular}
    % \vspace{-2mm}
    \caption{Rnd-effector's reaction force in the simulation.}
    \label{fig:sim_force}
    % \vspace{-2mm}
\end{figure}
\begin{figure}[t]
% \vspace{2mm}
    \begin{tabular}{cc}
     \begin{minipage}[t]{0.48\hsize}
         \centering
         \includegraphics[width=\linewidth]{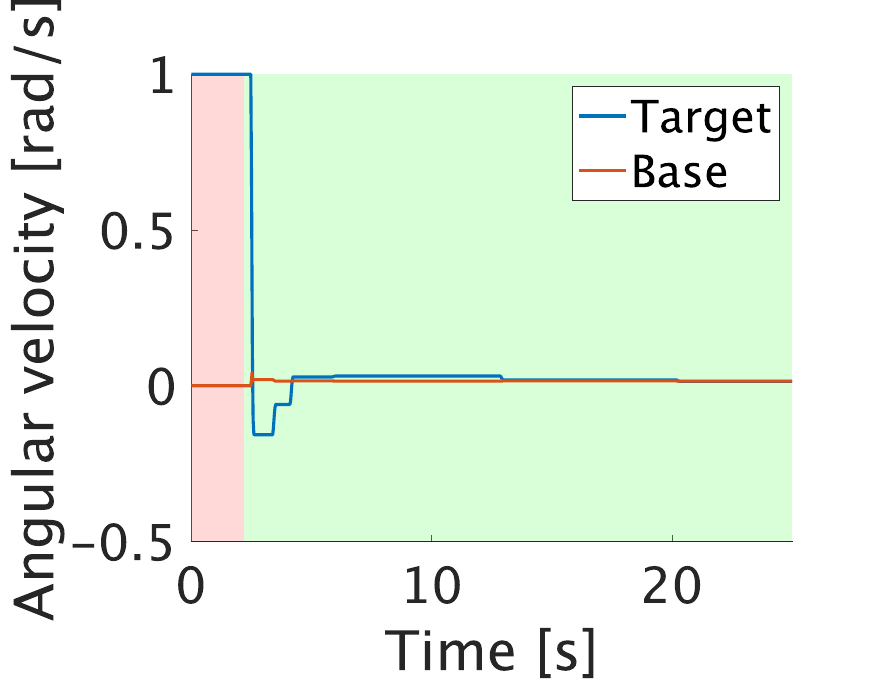}
         \subcaption{Direct caging.}
         \label{fig:sim_dir_w}
     \end{minipage}&  
      \begin{minipage}[t]{0.48\hsize}
         \centering
         \includegraphics[width=\linewidth]{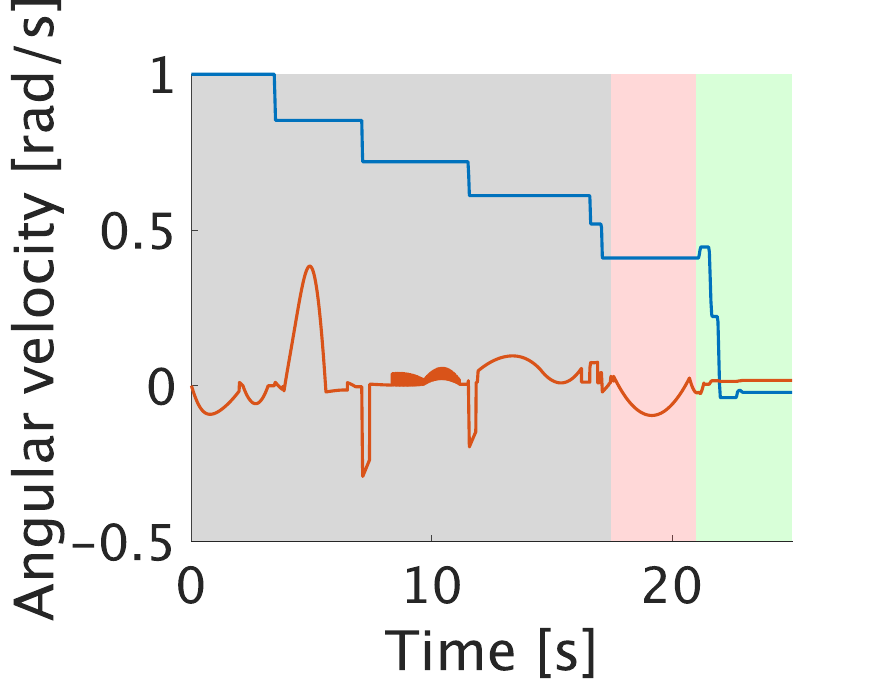}
         \subcaption{Detumbling and caging.}
         \label{fig:sim_mul_w}
     \end{minipage}\\
    \end{tabular}
    % \vspace{-2mm}
    \caption{Target and base angular velocities fluctuation in the simulation.}
    \label{fig:sim_target_w}
\vspace{-3mm}
\end{figure}
\begin{figure}[t]
% \vspace{2mm}
    \begin{tabular}{cc}
     \begin{minipage}[t]{0.48\hsize}
         \centering
         \includegraphics[width=.95\linewidth]{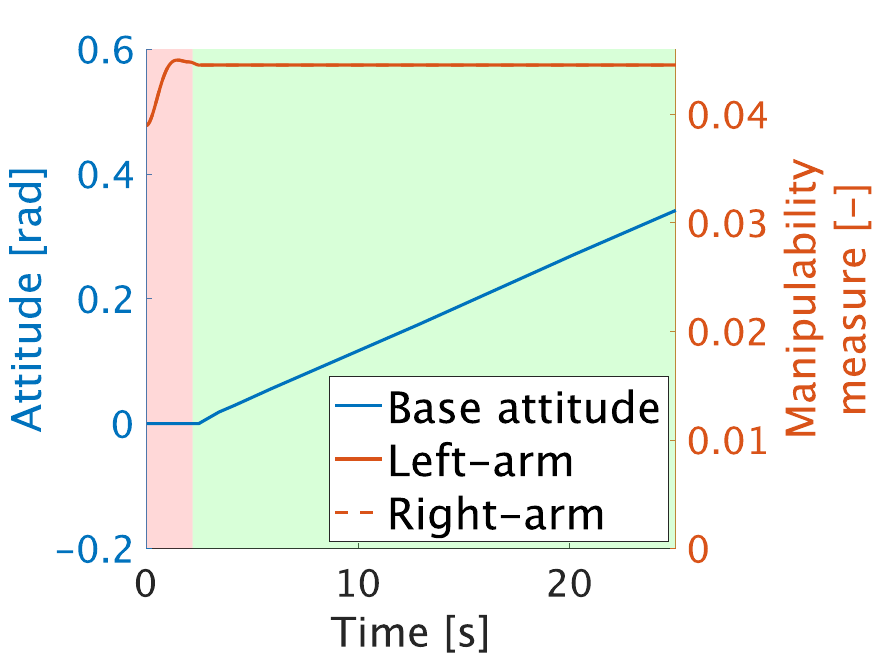}
         \subcaption{Direct caging.}
         \label{fig:sim_dir_stab}
     \end{minipage}&  
      \begin{minipage}[t]{0.48\hsize}
         \centering
         \includegraphics[width=.95\linewidth]{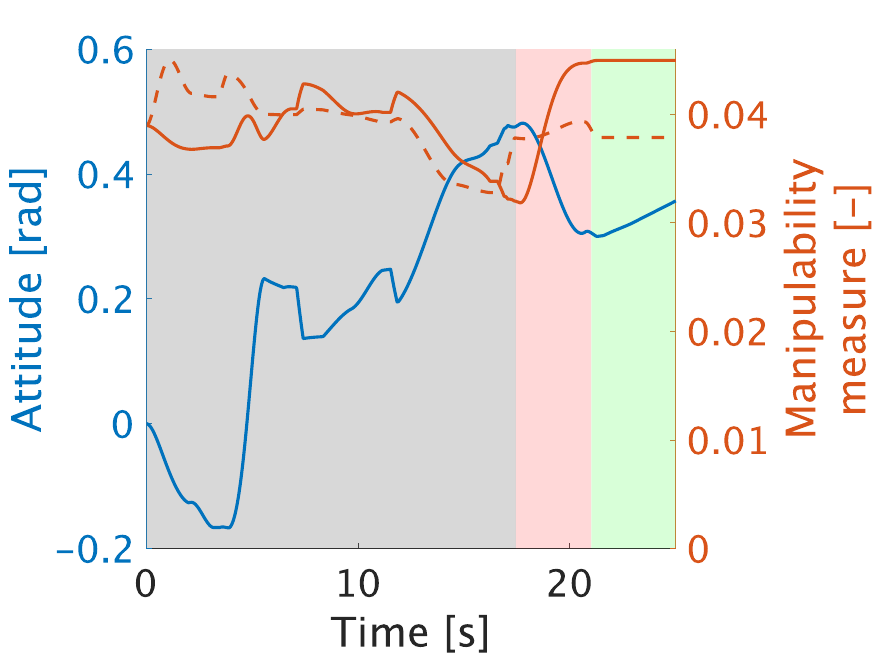}
         \subcaption{Detumbling and caging.}
         \label{fig:sim_mul_stab}
     \end{minipage}\\
    \end{tabular}
    % \vspace{-2mm}
    \caption{Base attitude and measure of manipulability in the simulation.}
    \label{fig:sim_stab}
\vspace{-3mm}
\end{figure}

\begin{table}[b]
\vspace{-4mm}
\caption{ROBOT PARAMETERS.}
\vspace{-2mm}
\centering{
 \scalebox{.78}{
 \begin{tabular}{ccccc}
 \hline
  & Mass [kg] & Inertia [kg$\cdot\mathrm{m}^2$] & Length [m] & COG [m]\\
 \hline
 Base & 8.31 & 0.135 & \begin{tabular}{c} $l_a=0.320$\\$l_b=0.160$\end{tabular} & 0.0761\\
 Link L${}_1$, R${}_1$ & 0.633 & $2.55\times 10^{-3}$ & 0.250 & 0.229\\
 Link L${}_2$, R${}_2$ & 0.647 & $1.19\times 10^{-3}$ & 0.175 & 0.162\\
 Link L${}_3$, R${}_3$ & 0.207 & $5.52\times 10^{-4}$ & \begin{tabular}{c}$l_c=0.137$\\$l_d=0.110$\\ $d=0.030$\end{tabular} & 0.0631\\
 Target& 2.35 & $9.33\times 10^{-3}$ & $l_t=0.150$ & 0.000\\
 \hline
 \end{tabular}}}
 \label{tab:robo_param}
 \vspace{-0mm}
\end{table}

\begin{table}[b]
% \vspace{-2mm}
\caption{CONTROL AND CONTACT PARAMETERS.}
\vspace{-2mm}
\centering
 \scalebox{.9}{
 \begin{tabular}{cc}
 \hline
 Parameter & Value\\
 \hline
 $k_\mathrm{p}$, $c_\mathrm{p}$, $\mu$ & 800 [N/m], 2 [N$\cdot$s/m], 0.1\\
 % $m_\mathrm{im}$, $d_\mathrm{im}$, $k_\mathrm{im}$ & 0.05 [kg], 0.3 [N$\cdot$s/m], 0.05 [N/m]\\
 $\alpha$, $\beta, \gamma$ & 0.8, 0.175 [rad],  0.175 [rad]\\
 \hline
 \end{tabular}} 
 \label{tab:cont_param}
 \vspace{-0mm}
\end{table}

% \section{VALIDATION BY SIMULATION}
\section{PARAMETRIC ANALYSIS SIMULATION}
The effect of the impedance parameters was substantial during the detumbling phase, underscoring the significance of their careful arrangement. Owing to the inherent challenges in theoretically determining optimal parameters, parametric analysis is an effective approach. Using mathematical simulations, the proposed detumbling technique was scrutinized under varying impedance parameters, namely, $\vecb{M}{im}, \vecb{D}{im}, \vecb{K}{im}$, to identify suitable parameter values.

\subsection{Condition}
The contact force was calculated employing the spring-damper model using the virtual deflection amount $\delta$ as follows, where $F_\mathrm{n} \in \vecu{R}{1}, F_\mathrm{t} \in \vecu{R}{1}$ are the normal and tangential components of the contact force, respectively. In addition, $k_\mathrm{p}, c_\mathrm{p}$, and $\mu$ denote the elastic modulus, damping coefficient, and friction coefficient, respectively. Furthermore, $v_{py}$ is the velocity of the end effector relative to the target in the interview direction.
\begin{equation}
   F_\mathrm{n} = (k_\mathrm{p} \delta + c_\mathrm{p} \dot{\delta})
\end{equation}
\begin{equation}
     F_\mathrm{t} = -\mathrm{sign}(v_{py}) \mu |F_\mathrm{n}| 
\end{equation}

The robot and control parameters used in the simulation are listed in Tables \tab{tab:robo_param} and \tab{tab:cont_param}, respectively. In this study, these parameters were set constant. The robot parameters followed those of the robot in \fig{fig:dar_new}, which were used in the experimental validation. Notably, the center of gravity (COG) was expressed by the $y$ component of the COG position of each link in the frame that was fixed to itself. Each link was symmetrical along the $x$ direction. Furthermore, $\vecb{M}{im}, \vecb{D}{im}, \vecb{K}{im}$ are defined as diagonal matrices represented by $\vecb{M}{im}=m_\mathrm{im}\vecu{E}{6}$, $\vecb{D}{im}=d_\mathrm{im}\vecu{E}{6}$, $\vecb{K}{im}=k_\mathrm{im}\vecu{E}{6}$, respectively. The threshold angular velocity for detumbling was set to \SI{0.5}{rad/s}. In addition to these parameters, trajectory feedback gains such as $\vecb{K}{p}$, $\vecb{K}{i}$, and $\vecb{K}{d}$ were set to $\vecb{0}{}$ because feedforward control was sufficiently accurate in the simulation. 

\begin{table*}[]
\caption{PARAMETRIC ANALYSIS RESULTS.}
    \centering
    \begin{tabular}{cc}
  (a) $\omega_\mathrm{0}=1.0$, $m_\mathrm{im}=0.01$ & (b) $\omega_\mathrm{0}=1.0$, $m_\mathrm{im}=0.05$
 \\
    \scalebox{.63}{
 \begin{tabular}{c|ccccccccccc}
 \hline
$k_\mathrm{im} \backslash d_\mathrm{im}$ & 0 & 1.5 & 3.0 & 4.5 & 6.0 & 7.5 & 9.0 & 10.5 & 12 & 13.5 & 15.0\\
\hline
0 & \cellcolor{myyellow}{$\checkmark    $} & \cellcolor{myyellow}{$\checkmark    $} & \cellcolor{myyellow}{$\checkmark    $} & \cellcolor{myred}{$\times          $} & \cellcolor{myred}{$\times          $} & \cellcolor{myred}{$\times          $} & \cellcolor{myred}{$\times          $} & \cellcolor{myred}{$\times          $} & \cellcolor{myred}{$\times          $} & \cellcolor{myred}{$\times          $} & \cellcolor{myred}{$\times          $}  \\
10 & \cellcolor{mylightgreen}{$\checkmark    $} & \cellcolor{mygreen}{$\checkmark    $} & \cellcolor{mygreen}{$\checkmark    $} & \cellcolor{mygreen}{$\checkmark    $} & \cellcolor{myred}{$\times          $} & \cellcolor{myred}{$\times          $} & \cellcolor{myred}{$\times          $} & \cellcolor{myred}{$\times          $} & \cellcolor{myred}{$\times          $} & \cellcolor{myred}{$\times          $} & \cellcolor{myred}{$\times          $}  \\
20 & \cellcolor{myred}{$\times          $} & \cellcolor{mygreen}{$\checkmark    $} & \cellcolor{mygreen}{$\checkmark    $} & \cellcolor{mygreen}{$\checkmark    $} & \cellcolor{mygreen}{$\checkmark    $} & \cellcolor{myred}{$\times          $} & \cellcolor{myred}{$\times          $} & \cellcolor{myred}{$\times          $} & \cellcolor{myred}{$\times          $} & \cellcolor{myred}{$\times          $} & \cellcolor{myred}{$\times          $}\\
30 & \cellcolor{myred}{$\times          $} & \cellcolor{mygreen}{$\checkmark    $} & \cellcolor{mygreen}{$\checkmark    $} & \cellcolor{mygreen}{$\checkmark    $} & \cellcolor{mygreen}{$\checkmark    $} & \cellcolor{mygreen}{$\checkmark    $} & \cellcolor{myred}{$\times          $} & \cellcolor{myred}{$\times          $} & \cellcolor{myred}{$\times          $} & \cellcolor{myred}{$\times          $} & \cellcolor{myred}{$\times          $} \\
40 & \cellcolor{myred}{$\times          $} & \cellcolor{myred}{$\times          $} & \cellcolor{mygreen}{$\checkmark    $} & \cellcolor{mygreen}{$\checkmark    $} & \cellcolor{mygreen}{$\checkmark    $} & \cellcolor{mygreen}{$\checkmark    $} & \cellcolor{mygreen}{$\checkmark    $} & \cellcolor{myred}{$\times          $} & \cellcolor{myred}{$\times          $} & \cellcolor{myred}{$\times          $} & \cellcolor{myred}{$\times          $}  \\
50 & \cellcolor{myred}{$\times          $} & \cellcolor{myred}{$\times          $} & \cellcolor{mygreen}{$\checkmark    $} & \cellcolor{mygreen}{$\checkmark    $} & \cellcolor{mygreen}{$\checkmark    $} & \cellcolor{mygreen}{$\checkmark    $} & \cellcolor{mygreen}{$\checkmark    $} & \cellcolor{mygreen}{$\checkmark    $} & \cellcolor{myred}{$\times          $} & \cellcolor{myred}{$\times          $} & \cellcolor{myred}{$\times          $} \\
60 & \cellcolor{myred}{$\times          $} & \cellcolor{myred}{$\times          $} & \cellcolor{mygreen}{$\checkmark    $} & \cellcolor{mygreen}{$\checkmark    $} & \cellcolor{mygreen}{$\checkmark    $} & \cellcolor{mygreen}{$\checkmark    $} & \cellcolor{mygreen}{$\checkmark    $} & \cellcolor{mygreen}{$\checkmark    $} & \cellcolor{myred}{$\times          $} & \cellcolor{myred}{$\times          $} & \cellcolor{myred}{$\times          $} \\
70 & \cellcolor{myred}{$\times          $} & \cellcolor{myred}{$\times          $} & \cellcolor{mygreen}{$\checkmark    $} & \cellcolor{mygreen}{$\checkmark    $} & \cellcolor{mygreen}{$\checkmark    $} & \cellcolor{mygreen}{$\checkmark    $} & \cellcolor{mygreen}{$\checkmark    $} & \cellcolor{mygreen}{$\checkmark    $} & \cellcolor{mygreen}{$\checkmark    $} & \cellcolor{myred}{$\times          $} & \cellcolor{myred}{$\times          $} \\
80 & \cellcolor{myred}{$\times          $} & \cellcolor{myred}{$\times          $} & \cellcolor{myyellow}{$\checkmark    $} & \cellcolor{mygreen}{$\checkmark    $} & \cellcolor{mygreen}{$\checkmark    $} & \cellcolor{mygreen}{$\checkmark    $} & \cellcolor{mygreen}{$\checkmark    $} & \cellcolor{mygreen}{$\checkmark    $} & \cellcolor{mygreen}{$\checkmark    $} & \cellcolor{mygreen}{$\checkmark    $} & \cellcolor{myred}{$\times          $}\\
90 & \cellcolor{myred}{$\times          $} & \cellcolor{myred}{$\times          $} & \cellcolor{myred}{$\times          $} & \cellcolor{mygreen}{$\checkmark    $} & \cellcolor{mygreen}{$\checkmark    $} & \cellcolor{mygreen}{$\checkmark    $} & \cellcolor{mygreen}{$\checkmark    $} & \cellcolor{mygreen}{$\checkmark    $} & \cellcolor{mygreen}{$\checkmark    $} & \cellcolor{mygreen}{$\checkmark    $} & \cellcolor{myred}{$\times          $} \\
100 & \cellcolor{myred}{$\times          $} & \cellcolor{myred}{$\times          $} & \cellcolor{myred}{$\times          $} & \cellcolor{mygreen}{$\checkmark    $} & \cellcolor{mygreen}{$\checkmark    $} & \cellcolor{mygreen}{$\checkmark    $} & \cellcolor{mygreen}{$\checkmark    $} & \cellcolor{mygreen}{$\checkmark    $} & \cellcolor{mygreen}{$\checkmark    $} & \cellcolor{mygreen}{$\checkmark    $} & \cellcolor{mygreen}{$\checkmark    $}  \\
 \hline
 \end{tabular}}
 &  
 \scalebox{.63}{
 \begin{tabular}{c|ccccccccccc}
 \hline
$k_\mathrm{im} \backslash d_\mathrm{im}$ & 0 & 1.5 & 3.0 & 4.5 & 6.0 & 7.5 & 9.0 & 10.5 & 12 & 13.5 & 15.0\\
\hline
0 & \cellcolor{myyellow}{$\checkmark    $} & \cellcolor{mylightgreen}{$\checkmark    $} & \cellcolor{myyellow}{$\checkmark    $} & \cellcolor{myred}{$\times    $} & \cellcolor{myred}{$\times    $} & \cellcolor{myred}{$\times    $} & \cellcolor{myred}{$\times    $} & \cellcolor{myred}{$\times    $} & \cellcolor{myred}{$\times    $} & \cellcolor{myred}{$\times    $} & \cellcolor{myred}{$\times    $} \\
10 & \cellcolor{mylightgreen}{$\checkmark    $} & \cellcolor{mygreen}{$\checkmark    $} & \cellcolor{mygreen}{$\checkmark    $} & \cellcolor{mygreen}{$\checkmark    $} & \cellcolor{myred}{$\times    $} & \cellcolor{myred}{$\times    $} & \cellcolor{myred}{$\times    $} & \cellcolor{myred}{$\times    $} & \cellcolor{myred}{$\times    $} & \cellcolor{myred}{$\times    $} & \cellcolor{myred}{$\times    $} \\
20 & \cellcolor{myred}{$\times    $} & \cellcolor{mygreen}{$\checkmark    $} & \cellcolor{mygreen}{$\checkmark    $} & \cellcolor{mygreen}{$\checkmark    $} & \cellcolor{mygreen}{$\checkmark    $} & \cellcolor{myred}{$\times    $} & \cellcolor{myred}{$\times    $} & \cellcolor{myred}{$\times    $} & \cellcolor{myred}{$\times    $} & \cellcolor{myred}{$\times    $} & \cellcolor{myred}{$\times    $}  \\
30 & \cellcolor{myred}{$\times    $} & \cellcolor{mygreen}{$\checkmark    $} & \cellcolor{mygreen}{$\checkmark    $} & \cellcolor{mygreen}{$\checkmark    $} & \cellcolor{mygreen}{$\checkmark    $} & \cellcolor{mygreen}{$\checkmark    $} & \cellcolor{myred}{$\times    $} & \cellcolor{myred}{$\times    $} & \cellcolor{myred}{$\times    $} & \cellcolor{myred}{$\times    $} & \cellcolor{myred}{$\times    $} \\
40 & \cellcolor{myred}{$\times    $} & \cellcolor{myred}{$\times    $} & \cellcolor{mygreen}{$\checkmark    $} & \cellcolor{mygreen}{$\checkmark    $} & \cellcolor{mygreen}{$\checkmark    $} & \cellcolor{mygreen}{$\checkmark    $} & \cellcolor{mygreen}{$\checkmark    $} & \cellcolor{myred}{$\times    $} & \cellcolor{myred}{$\times    $} & \cellcolor{myred}{$\times    $} & \cellcolor{myred}{$\times    $} \\
50 & \cellcolor{myred}{$\times    $} & \cellcolor{myred}{$\times    $} & \cellcolor{mygreen}{$\checkmark    $} & \cellcolor{mygreen}{$\checkmark    $} & \cellcolor{mygreen}{$\checkmark    $} & \cellcolor{mygreen}{$\checkmark    $} & \cellcolor{mygreen}{$\checkmark    $} & \cellcolor{mygreen}{$\checkmark    $} & \cellcolor{myred}{$\times    $} & \cellcolor{myred}{$\times    $} & \cellcolor{myred}{$\times    $} \\
60 & \cellcolor{myred}{$\times    $} & \cellcolor{myred}{$\times    $} & \cellcolor{mygreen}{$\checkmark    $} & \cellcolor{mygreen}{$\checkmark    $} & \cellcolor{mygreen}{$\checkmark    $} & \cellcolor{mygreen}{$\checkmark    $} & \cellcolor{mygreen}{$\checkmark    $} & \cellcolor{mygreen}{$\checkmark    $} & \cellcolor{myred}{$\times    $} & \cellcolor{myred}{$\times    $} & \cellcolor{myred}{$\times    $} \\
70 & \cellcolor{myred}{$\times    $} & \cellcolor{myred}{$\times    $} & \cellcolor{mygreen}{$\checkmark    $} & \cellcolor{mygreen}{$\checkmark    $} & \cellcolor{mygreen}{$\checkmark    $} & \cellcolor{mygreen}{$\checkmark    $} & \cellcolor{mygreen}{$\checkmark    $} & \cellcolor{mygreen}{$\checkmark    $} & \cellcolor{mygreen}{$\checkmark    $} & \cellcolor{myred}{$\times    $} & \cellcolor{myred}{$\times    $} \\
80 & \cellcolor{myred}{$\times    $} & \cellcolor{myred}{$\times    $} & \cellcolor{mygreen}{$\checkmark    $} & \cellcolor{mygreen}{$\checkmark    $} & \cellcolor{mygreen}{$\checkmark    $} & \cellcolor{mygreen}{$\checkmark    $} & \cellcolor{mygreen}{$\checkmark    $} & \cellcolor{mygreen}{$\checkmark    $} & \cellcolor{mygreen}{$\checkmark    $} & \cellcolor{myred}{$\times    $} & \cellcolor{myred}{$\times    $} \\
90 & \cellcolor{myred}{$\times    $} & \cellcolor{myred}{$\times    $} & \cellcolor{myyellow}{$\checkmark    $} & \cellcolor{mygreen}{$\checkmark    $} & \cellcolor{mygreen}{$\checkmark    $} & \cellcolor{mygreen}{$\checkmark    $} & \cellcolor{mygreen}{$\checkmark    $} & \cellcolor{mygreen}{$\checkmark    $} & \cellcolor{mygreen}{$\checkmark    $} & \cellcolor{mygreen}{$\checkmark    $} & \cellcolor{myred}{$\times    $} \\
100 & \cellcolor{myred}{$\times    $} & \cellcolor{myred}{$\times    $} & \cellcolor{myred}{$\times    $} & \cellcolor{mygreen}{$\checkmark    $} & \cellcolor{mygreen}{$\checkmark    $} & \cellcolor{mygreen}{$\checkmark    $} & \cellcolor{mygreen}{$\checkmark    $} & \cellcolor{mygreen}{$\checkmark    $} & \cellcolor{mygreen}{$\checkmark    $} & \cellcolor{mygreen}{$\checkmark    $} & \cellcolor{mygreen}{$\checkmark    $} \\
 \hline
 \end{tabular}}
 
 \\
 \\
   (c) $\omega_\mathrm{0}=1.0$, $m_\mathrm{im}=0.1$ & (d) $\omega_\mathrm{0}=1.0$, $m_\mathrm{im}=0.5$
 \\

 \scalebox{.63}{
 \begin{tabular}{c|ccccccccccc}
 \hline
$k_\mathrm{im} \backslash d_\mathrm{im}$ & 0 & 1.5 & 3.0 & 4.5 & 6.0 & 7.5 & 9.0 & 10.5 & 12 & 13.5 & 15.0\\
\hline
0 & \cellcolor{myyellow}{$\checkmark    $} & \cellcolor{mylightgreen}{$\checkmark    $} & \cellcolor{myyellow}{$\checkmark    $} & \cellcolor{myred}{$\times    $} & \cellcolor{myred}{$\times    $} & \cellcolor{myred}{$\times    $} & \cellcolor{myred}{$\times    $} & \cellcolor{myred}{$\times    $} & \cellcolor{myred}{$\times    $} & \cellcolor{myred}{$\times    $} & \cellcolor{myred}{$\times    $} \\
10 & \cellcolor{mygreen}{$\checkmark    $} & \cellcolor{mygreen}{$\checkmark    $} & \cellcolor{mygreen}{$\checkmark    $} & \cellcolor{mygreen}{$\checkmark    $} & \cellcolor{myred}{$\times    $} & \cellcolor{myred}{$\times    $} & \cellcolor{myred}{$\times    $} & \cellcolor{myred}{$\times    $} & \cellcolor{myred}{$\times    $} & \cellcolor{myred}{$\times    $} & \cellcolor{myred}{$\times    $}  \\
20 & \cellcolor{myred}{$\times    $} & \cellcolor{mygreen}{$\checkmark    $} & \cellcolor{mygreen}{$\checkmark    $} & \cellcolor{mygreen}{$\checkmark    $} & \cellcolor{mygreen}{$\checkmark    $} & \cellcolor{myred}{$\times    $} & \cellcolor{myred}{$\times    $} & \cellcolor{myred}{$\times    $} & \cellcolor{myred}{$\times    $} & \cellcolor{myred}{$\times    $} & \cellcolor{myred}{$\times    $} \\
30 & \cellcolor{myred}{$\times    $} & \cellcolor{mygreen}{$\checkmark    $} & \cellcolor{mygreen}{$\checkmark    $} & \cellcolor{mygreen}{$\checkmark    $} & \cellcolor{mygreen}{$\checkmark    $} & \cellcolor{mygreen}{$\checkmark    $} & \cellcolor{myred}{$\times    $} & \cellcolor{myred}{$\times    $} & \cellcolor{myred}{$\times    $} & \cellcolor{myred}{$\times    $} & \cellcolor{myred}{$\times    $} \\
40 & \cellcolor{myred}{$\times    $} & \cellcolor{mygreen}{$\checkmark    $} & \cellcolor{mygreen}{$\checkmark    $} & \cellcolor{mygreen}{$\checkmark    $} & \cellcolor{mygreen}{$\checkmark    $} & \cellcolor{mygreen}{$\checkmark    $} & \cellcolor{mygreen}{$\checkmark    $} & \cellcolor{myred}{$\times    $} & \cellcolor{myred}{$\times    $} & \cellcolor{myred}{$\times    $} & \cellcolor{myred}{$\times    $}  \\
50 & \cellcolor{myred}{$\times    $} & \cellcolor{myred}{$\times    $} & \cellcolor{mygreen}{$\checkmark    $} & \cellcolor{mygreen}{$\checkmark    $} & \cellcolor{mygreen}{$\checkmark    $} & \cellcolor{mygreen}{$\checkmark    $} & \cellcolor{mygreen}{$\checkmark    $} & \cellcolor{myred}{$\times    $} & \cellcolor{myred}{$\times    $} & \cellcolor{myred}{$\times    $} & \cellcolor{myred}{$\times    $} \\
60 & \cellcolor{myred}{$\times    $} & \cellcolor{myred}{$\times    $} & \cellcolor{mygreen}{$\checkmark    $} & \cellcolor{mygreen}{$\checkmark    $} & \cellcolor{mygreen}{$\checkmark    $} & \cellcolor{mygreen}{$\checkmark    $} & \cellcolor{mygreen}{$\checkmark    $} & \cellcolor{mygreen}{$\checkmark    $} & \cellcolor{myred}{$\times    $} & \cellcolor{myred}{$\times    $} & \cellcolor{myred}{$\times    $}  \\
70 & \cellcolor{myred}{$\times    $} & \cellcolor{myred}{$\times    $} & \cellcolor{mygreen}{$\checkmark    $} & \cellcolor{mygreen}{$\checkmark    $} & \cellcolor{mygreen}{$\checkmark    $} & \cellcolor{mygreen}{$\checkmark    $} & \cellcolor{mygreen}{$\checkmark    $} & \cellcolor{mygreen}{$\checkmark    $} & \cellcolor{mygreen}{$\checkmark    $} & \cellcolor{myred}{$\times    $} & \cellcolor{myred}{$\times    $}\\
80 & \cellcolor{myred}{$\times    $} & \cellcolor{myred}{$\times    $} & \cellcolor{mygreen}{$\checkmark    $} & \cellcolor{mygreen}{$\checkmark    $} & \cellcolor{mygreen}{$\checkmark    $} & \cellcolor{mygreen}{$\checkmark    $} & \cellcolor{mygreen}{$\checkmark    $} & \cellcolor{mygreen}{$\checkmark    $} & \cellcolor{mygreen}{$\checkmark    $} & \cellcolor{myred}{$\times    $} & \cellcolor{myred}{$\times    $} \\
90 & \cellcolor{myred}{$\times    $} & \cellcolor{myred}{$\times    $} & \cellcolor{mygreen}{$\checkmark    $} & \cellcolor{mygreen}{$\checkmark    $} & \cellcolor{mygreen}{$\checkmark    $} & \cellcolor{mygreen}{$\checkmark    $} & \cellcolor{mygreen}{$\checkmark    $} & \cellcolor{mygreen}{$\checkmark    $} & \cellcolor{mygreen}{$\checkmark    $} & \cellcolor{mygreen}{$\checkmark    $} & \cellcolor{myred}{$\times    $} \\
100 & \cellcolor{myred}{$\times    $} & \cellcolor{myred}{$\times    $} & \cellcolor{myred}{$\times    $} & \cellcolor{mygreen}{$\checkmark    $} & \cellcolor{mygreen}{$\checkmark    $} & \cellcolor{mygreen}{$\checkmark    $} & \cellcolor{mygreen}{$\checkmark    $} & \cellcolor{mygreen}{$\checkmark    $} & \cellcolor{mygreen}{$\checkmark    $} & \cellcolor{mygreen}{$\checkmark    $} & \cellcolor{myred}{$\times    $} \\
 \hline
 \end{tabular}}
 
    &

\scalebox{.63}{
 \begin{tabular}{c|ccccccccccc}
 \hline
$k_\mathrm{im} \backslash d_\mathrm{im}$ & 0 & 1.5 & 3.0 & 4.5 & 6.0 & 7.5 & 9.0 & 10.5 & 12 & 13.5 & 15.0\\
\hline
0 & \cellcolor{myyellow}{$\checkmark    $} & \cellcolor{myyellow}{$\checkmark    $} & \cellcolor{myred}{$\times          $} & \cellcolor{myred}{$\times          $} & \cellcolor{myred}{$\times          $} & \cellcolor{myred}{$\times          $} & \cellcolor{myred}{$\times          $} & \cellcolor{myred}{$\times          $} & \cellcolor{myred}{$\times          $} & \cellcolor{myred}{$\times          $} & \cellcolor{myred}{$\times          $} \\
10 & \cellcolor{mylightgreen}{$\checkmark    $} & \cellcolor{myred}{$\times          $} & \cellcolor{myred}{$\times          $} & \cellcolor{myred}{$\times          $} & \cellcolor{myred}{$\times          $} & \cellcolor{myred}{$\times          $} & \cellcolor{myred}{$\times          $} & \cellcolor{myred}{$\times          $} & \cellcolor{myred}{$\times          $} & \cellcolor{myred}{$\times          $} & \cellcolor{myred}{$\times          $}\\
20 & \cellcolor{mygreen}{$\checkmark    $} & \cellcolor{mygreen}{$\checkmark    $} & \cellcolor{mygreen}{$\checkmark    $} & \cellcolor{mygreen}{$\checkmark    $} & \cellcolor{myred}{$\times          $} & \cellcolor{myred}{$\times          $} & \cellcolor{myred}{$\times          $} & \cellcolor{myred}{$\times          $} & \cellcolor{myred}{$\times          $} & \cellcolor{myred}{$\times          $} & \cellcolor{myred}{$\times          $} \\
30 & \cellcolor{mygreen}{$\checkmark    $} & \cellcolor{mygreen}{$\checkmark    $} & \cellcolor{mygreen}{$\checkmark    $} & \cellcolor{mygreen}{$\checkmark    $} & \cellcolor{mygreen}{$\checkmark    $} & \cellcolor{mygreen}{$\checkmark    $} & \cellcolor{myred}{$\times          $} & \cellcolor{myred}{$\times          $} & \cellcolor{myred}{$\times          $} & \cellcolor{myred}{$\times          $} & \cellcolor{myred}{$\times          $} \\
40 & \cellcolor{mygreen}{$\checkmark    $} & \cellcolor{mygreen}{$\checkmark    $} & \cellcolor{mygreen}{$\checkmark    $} & \cellcolor{mygreen}{$\checkmark    $} & \cellcolor{mygreen}{$\checkmark    $} & \cellcolor{mygreen}{$\checkmark    $} & \cellcolor{myred}{$\times          $} & \cellcolor{myred}{$\times          $} & \cellcolor{myred}{$\times          $} & \cellcolor{myred}{$\times          $} & \cellcolor{myred}{$\times          $} \\
50 & \cellcolor{mygreen}{$\checkmark    $} & \cellcolor{mygreen}{$\checkmark    $} & \cellcolor{mygreen}{$\checkmark    $} & \cellcolor{mygreen}{$\checkmark    $} & \cellcolor{mygreen}{$\checkmark    $} & \cellcolor{mygreen}{$\checkmark    $} & \cellcolor{mygreen}{$\checkmark    $} & \cellcolor{myred}{$\times          $} & \cellcolor{myred}{$\times          $} & \cellcolor{myred}{$\times          $} & \cellcolor{myred}{$\times          $}  \\
60 & \cellcolor{mygreen}{$\checkmark    $} & \cellcolor{mygreen}{$\checkmark    $} & \cellcolor{mygreen}{$\checkmark    $} & \cellcolor{mygreen}{$\checkmark    $} & \cellcolor{mygreen}{$\checkmark    $} & \cellcolor{mygreen}{$\checkmark    $} & \cellcolor{mygreen}{$\checkmark    $} & \cellcolor{myred}{$\times          $} & \cellcolor{myred}{$\times          $} & \cellcolor{myred}{$\times          $} & \cellcolor{myred}{$\times          $}  \\
70 & \cellcolor{mygreen}{$\checkmark    $} & \cellcolor{mygreen}{$\checkmark    $} & \cellcolor{mygreen}{$\checkmark    $} & \cellcolor{mygreen}{$\checkmark    $} & \cellcolor{mygreen}{$\checkmark    $} & \cellcolor{mygreen}{$\checkmark    $} & \cellcolor{mygreen}{$\checkmark    $} & \cellcolor{myred}{$\times          $} & \cellcolor{myred}{$\times          $} & \cellcolor{myred}{$\times          $} & \cellcolor{myred}{$\times          $}\\
80 & \cellcolor{mygreen}{$\checkmark    $} & \cellcolor{mygreen}{$\checkmark    $} & \cellcolor{mygreen}{$\checkmark    $} & \cellcolor{mygreen}{$\checkmark    $} & \cellcolor{mygreen}{$\checkmark    $} & \cellcolor{mygreen}{$\checkmark    $} & \cellcolor{mygreen}{$\checkmark    $} & \cellcolor{mygreen}{$\checkmark    $} & \cellcolor{myred}{$\times          $} & \cellcolor{myred}{$\times          $} & \cellcolor{myred}{$\times          $}  \\
90 & \cellcolor{mygreen}{$\checkmark    $} & \cellcolor{mygreen}{$\checkmark    $} & \cellcolor{mygreen}{$\checkmark    $} & \cellcolor{mygreen}{$\checkmark    $} & \cellcolor{mygreen}{$\checkmark    $} & \cellcolor{mygreen}{$\checkmark    $} & \cellcolor{mygreen}{$\checkmark    $} & \cellcolor{mygreen}{$\checkmark    $} & \cellcolor{myred}{$\times          $} & \cellcolor{myred}{$\times          $} & \cellcolor{myred}{$\times          $}  \\
100 & \cellcolor{myred}{$\times          $} & \cellcolor{mygreen}{$\checkmark    $} & \cellcolor{mygreen}{$\checkmark    $} & \cellcolor{mygreen}{$\checkmark    $} & \cellcolor{mygreen}{$\checkmark    $} & \cellcolor{mygreen}{$\checkmark    $} & \cellcolor{mygreen}{$\checkmark    $} & \cellcolor{mygreen}{$\checkmark    $} & \cellcolor{mygreen}{$\checkmark    $} & \cellcolor{myred}{$\times          $} & \cellcolor{myred}{$\times          $}  \\
\hline
 \end{tabular}}
 \\

% \centering
% \scalebox{1}{
%  \begin{tabular}{c|cc}
%  \hline
% \cellcolor{mygreen}{$\checkmark    $} & Success:& Caging with the force decreased. \\
% \cellcolor{mylightgreen}{$\checkmark    $} & Success:& Caging without force reduction. \\
% \cellcolor{myyellow}{$\checkmark    $} & Success:& Caging with unplanned contacts. \\
%  \hline
%  \end{tabular}}
 
%  & 

% \centering
% \scalebox{1}{
%  \begin{tabular}{c|cc}
%  \hline
% \cellcolor{myorange}{$\times    $} & Failure:& Target ejection. \\
% \cellcolor{myred}{$\times    $} & Failure:& Collision on the base. \\
%  \hline
%  \end{tabular}}

    \end{tabular}
    \label{table:parametric}
\end{table*}

\begin{table}[]
    \centering
\centering
\scalebox{0.78}{
 \begin{tabular}{c|cc}
 \hline
\cellcolor{mygreen}{$\checkmark    $} & Success:& Caging with effective force reduction. \\
\cellcolor{mylightgreen}{$\checkmark    $} & Success:& Caging without force reduction. \\
\cellcolor{myyellow}{$\checkmark    $} & Success:& Caging with unplanned contacts. \\
\cellcolor{myorange}{$\times    $} & Failure:& Target ejection. \\
\cellcolor{myred}{$\times    $} & Failure:& Collision on the base. \\
 \hline
 \end{tabular}}
\end{table}

A parametric analysis was performed for the target angular velocity  $w_\mathrm{0}=\SI{1.0}{ras/s}$ by changing the impedance parameter as follows. In this study, the initial angular velocity of the target was not varied to focus on revealing the effect of each impedance parameter. Herein, $m_\mathrm{im}=0.01, 0.05, 0.1, \SI{0.5}{kg}$, $d_\mathrm{im}=0.0, 1.5, ... \SI{15.0}{N\cdot s/m}$ and $k_\mathrm{im}=0.0, 10, ... \SI{100}{N/m}$.

\subsection{Results}
The results of the parametric analysis are presented in Table \tab{table:parametric}. The symbols and colors in the tables indicate the results of the capture sequence. In the context of the capturing simulation, five distinct outcomes characterized the final results of a sequence. 1) Successful caging, where the maximum force applied was reduced compared to the direct caging scenario. 2) Successful caging with a maximum force exceeding that of direct caging. 3) Successful caging with unplanned contacts. 4) Failure marked by the ejection of the target. 5) Failure attributed to a collision between the robot base and target.

As a representative case of the analysis, the simulation results under the conditions $m_\mathrm{im}=0.10$, $d_\mathrm{im}=1.5$, $k_\mathrm{im}=10.0$ are presented in \fig{fig:sim_snapshots}, showing snapshots of the robot and target during the sequence, which were captured at the representative scene. \fig{fig:sim_force} presents a comparison of the force applied to the tip spheres of the end effector between direct caging and detumbling and caging by impedance-controlled repeated contacts. \fig{fig:sim_target_w} also shows the angular velocity of the target and robot base in the direct caging and repeated contacts utilized cases. \mbox{\fig{fig:sim_stab}} presents a comparison of the robot base orientation and measure of manipulability, which was defined as a volume of the manipulability ellipsoid related to the end effector's velocity, for the two cases.

\subsection{Discussion}
% Tables \ref{table:w3ma} to \ref{table:w1mc} demonstrate that when $\omega_\mathrm{0}=3.0$, the region of appropriate parameters is notably smaller compared to the case where $\omega_\mathrm{0}=1.0$. That means if the target angular velocity is larger, it becomes more difficult to find suitable parameters, and it is consistent with the fact that capturing becomes more challenging with the increasing angular velocity. Additionally, comparing the difference between \tab{table:w1ma} and \tab{table:w1md}, and the one between \tab{table:w3ma} and \tab{table:w3md}, it is indicated that the effect of virtual mass is more significant for large angular velocity than for small one. This is thought to be because when the angular velocity of the target is small, the impact at contact is small and the target is hard to be pushed away, so it continuously contacts the end effector after contact and is greatly affected by the damping term, while when the angular velocity of the target is large, the target can be bounced away immediately after contact, so the end effector must behave as if it is being bounced away by the target to avoid the target ejection, and the virtual mass effect is greater than other coefficients.

\fig{fig:sim_snapshots} shows that the robot realized successful detumbling and caging, as a representative case of the parametric simulation. \fig{fig:sim_force} indicates that the maximum force in this case sequence was reduced \SI{50}{\%}, from \SI{1.2}{N} to \SI{0.60}{N} compared with the direct caging case. Furthermore, \fig{fig:sim_target_w} shows the gradual mitigation of the target angular velocity through repeated contact. These results indicate that the target could be successfully detumbled with a relatively low impact, and its momentum exchange could be distributed multiple times using the proposed method. \mbox{\fig{fig:sim_stab}} indicates that the robot attitude was dynamically changing during the multiple contact-based detumbling sequences resulting in decreased manipulability compared with the direct caging case. However, the volume of manipulability ellipsoid was still beyond zero, implying that the robot is sufficiently far from the singularity.

\tab{table:parametric} shows that there was a certain area for the impedance parameters which was suitable for successful detumbling and capturing. These results indicated that the damping coefficient was more sensitive than the elastic coefficient. The tables also indicate that, for successful detumbling and caging, the damping coefficient must increase as the elastic coefficient increases. Moreover, the influence of the virtual mass parameter was relatively low compared to that of the other parameters.

% \begin{figure}[t]
% % \renewcommand{\baselinestretch}{0.6}
% % \vspace{2mm}
%   \centering
%   \includegraphics[width=0.8\linewidth]{fig/target.pdf}
%   \vspace{0mm}\caption{Target testbed. }\label{fig:target}
% % \vspace{2mm}
% \end{figure}

\section{EXPERIMENTAL VALIDATION}
\subsection{Condition}
\begin{figure}[t]
\vspace{5mm}
  \centering
  \includegraphics[width=.68\linewidth]{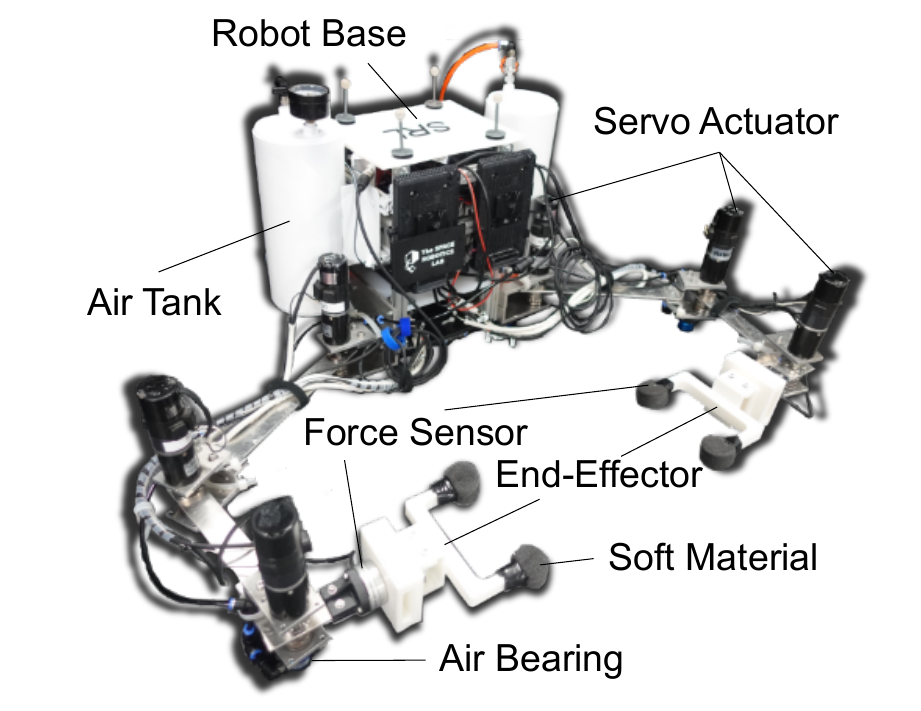}
  \caption{Dual-arm orbital robot testbed in SRL.}\label{fig:dar_new}
% \vspace{-5mm}
\end{figure}

\begin{figure}[t]
\vspace{5mm}
  \centering
  \includegraphics[width=0.75\linewidth]{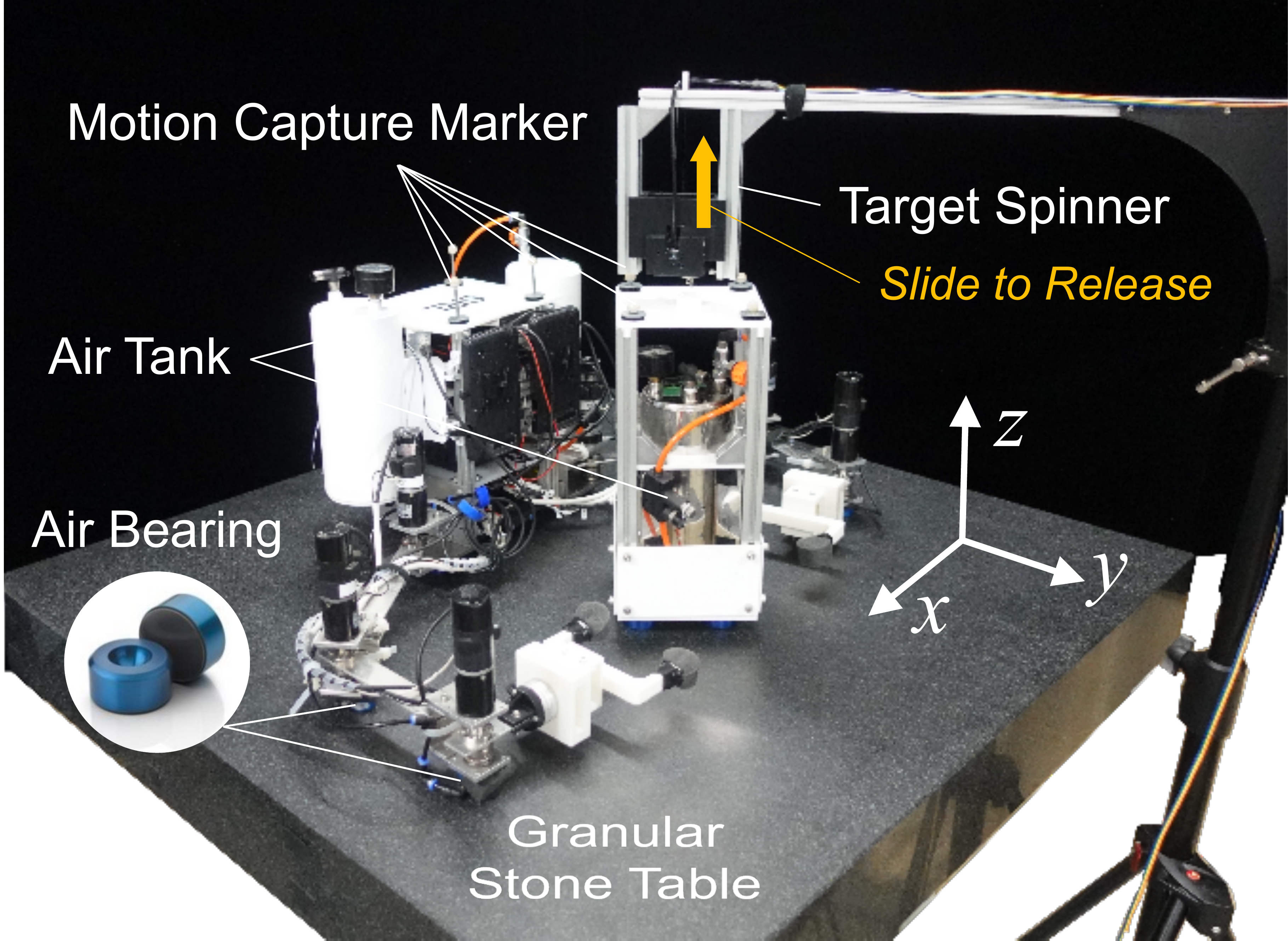}
  \vspace{0mm}\caption{Air floating microgravity experiment setup.}\label{fig:AFT}
% \vspace{-5mm}
\end{figure}

\fig{fig:dar_new} shows the robot used in the experiment. The robot was developed by the authors as a successor model to the SRL dual-arm robot. It was equipped with an onboard PC connected to motion capture via wifi; thus, it knew the target and its pose in time. In addition, there was a force sensor on each wrist part on the rear end of the end effector, which enabled the robot to conduct admittance control. The proposed sequence was computed on the robot online. The control parameters were the same as those used in the simulation, except that the trajectory feedback gains were as follows: ${\vecb{K}{p}=2.}5$, ${\vecb{K}{i}=0.10}$, ${\vecb{K}{d}=0.0}$. The impedance parameters were configured as follows: ${m_\mathrm{im}=0.10}$, ${d_\mathrm{im}=1.5}$, ${k_\mathrm{im}=10.0}$. These values were consistent with the representative simulation results shown in \fig{fig:sim_snapshots}, demonstrating their suitability for achieving successful detumbling through parametric analysis simulations.
The target and air-floating experimental environments used in the real-world validation are shown in \fig{fig:AFT}, where the two-dimensional microgravity conditions were emulated using a compressed air flow. The target mock-up was spined by a velocity-controlled motor and then released. 

\subsection{Results}
\begin{figure*}[!t]
\centering
    \includegraphics[width=.9\linewidth]{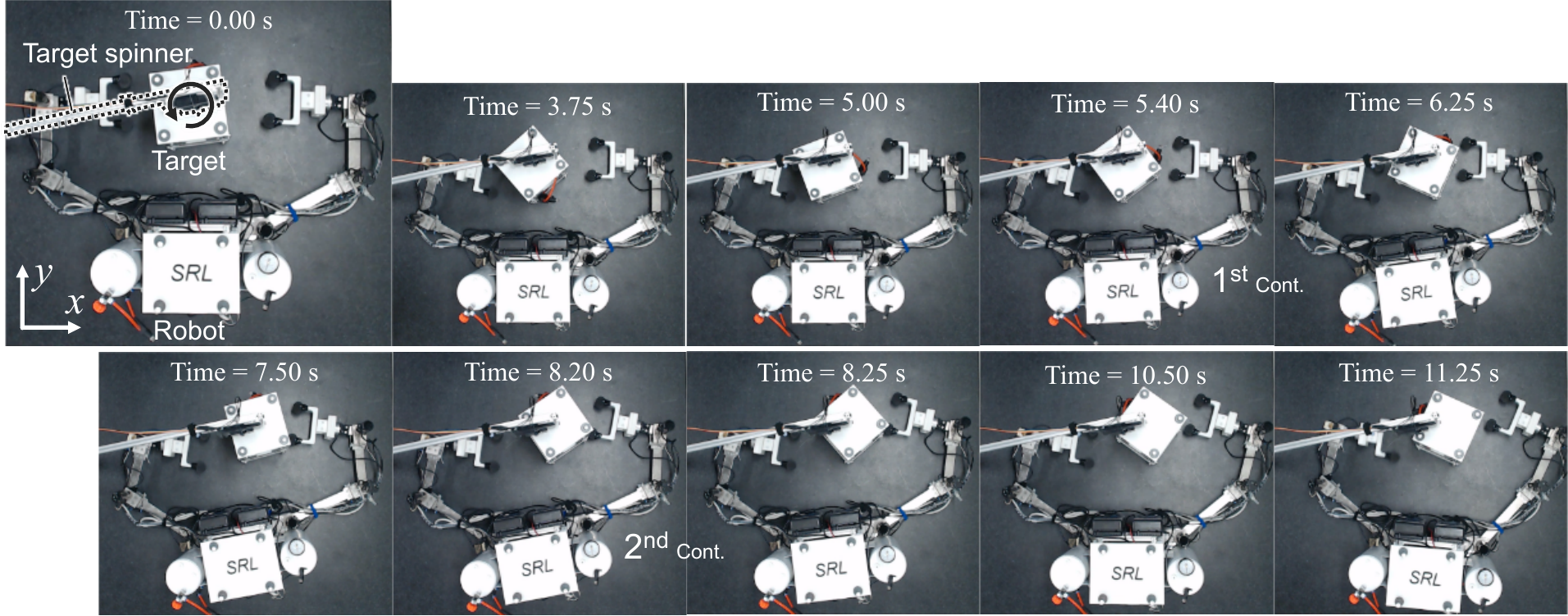}
    % \vspace{-2mm}
    \caption{Snapshots of the space debris capturing experiment.}
    \label{fig:exp_snapshots}
    % \vspace{-3mm}
\end{figure*}

\figs{fig:exp_snapshots}{fig:exp_stab} shows the experimental results.
Snapshots of the experiment are presented in \fig{fig:exp_snapshots}, where the dual-arm robot, target, and target spinner can be observed. Snapshots were captured in periodic and representative scenes. \fig{fig:exp_force} and \fig{fig:exp_target_w} show the force detected by the force sensor on each wrist and the target angular velocity obtained by the motion capture cameras, respectively. \mbox{\fig{fig:exp_stab}} shows the robot base attitude and measure of manipulability, or the volume of the manipulability ellipsoid.

\subsection{Discussion}
\begin{figure}[!t]
% \vspace{3mm}
    \begin{tabular}{cc}
     \begin{minipage}[ht]{0.48\hsize}
         \centering
         \includegraphics[height=0.75\linewidth]{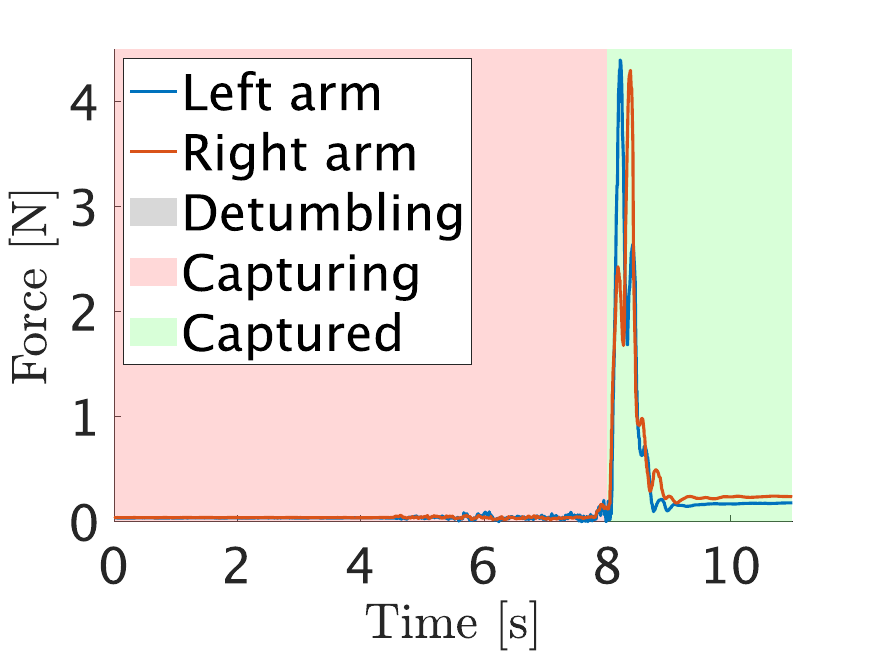}
         \subcaption{Direct caging.}
         \label{fig:exp_dir_f}
     \end{minipage}&  
      \begin{minipage}[ht]{0.48\hsize}
         \centering
         \includegraphics[height=0.75\linewidth]{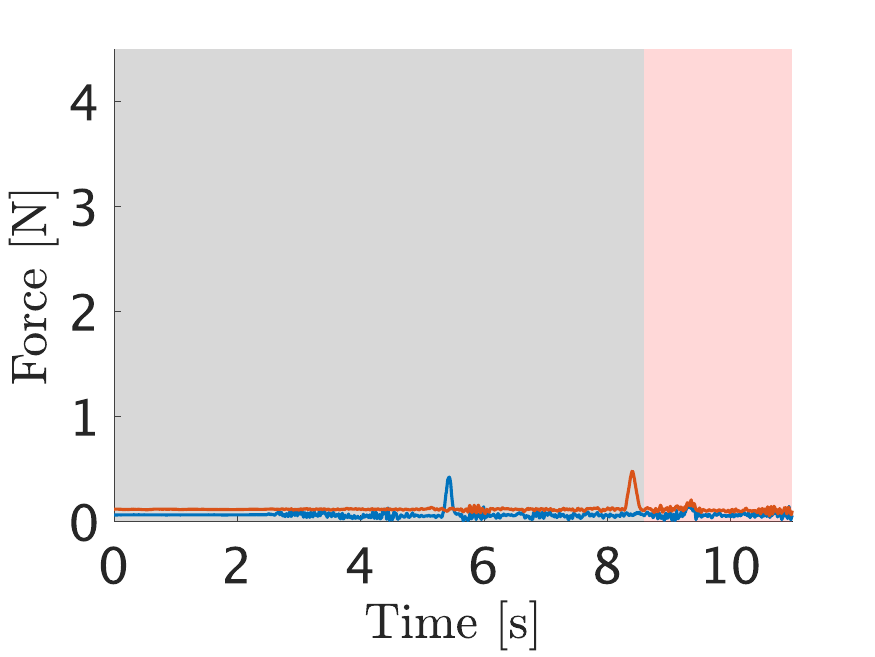}
         \subcaption{Detumbling and caging.}
         \label{fig:exp_mul_f}
     \end{minipage}\\
    \end{tabular}
    % \vspace{-2mm}
    \caption{End effector's reaction force in the experiment.}
    \label{fig:exp_force}
    \vspace{0mm}
\end{figure}

\begin{figure}[!t]
% \vspace{3mm}
    \begin{tabular}{cc}
     \begin{minipage}[t]{0.48\hsize}
         \centering
         \includegraphics[height=0.75\linewidth]{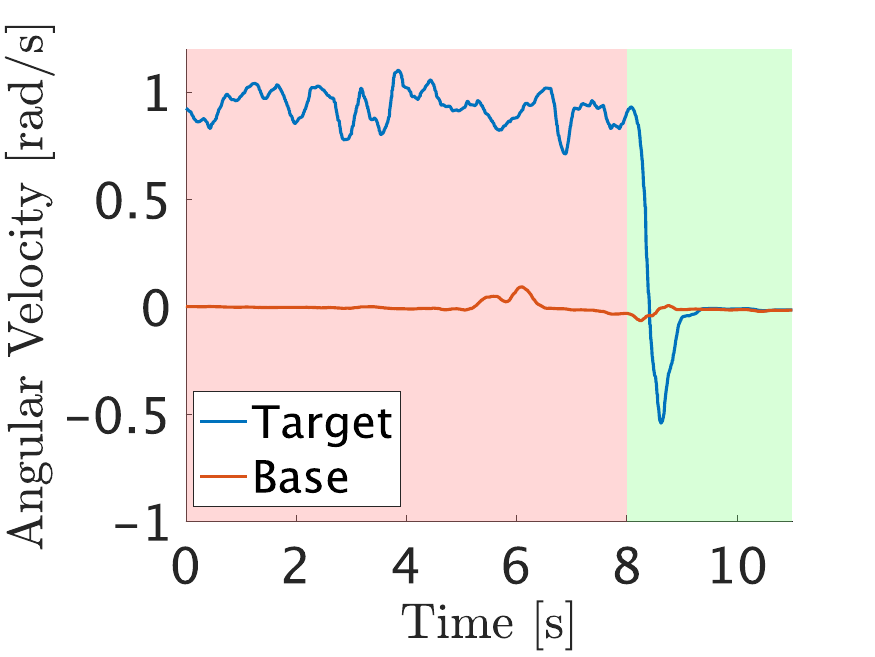}
         \subcaption{Direct caging.}
         \label{fig:exp_dir_w}
     \end{minipage}&  
      \begin{minipage}[t]{0.48\hsize}
         \centering
         \includegraphics[height=0.75\linewidth]{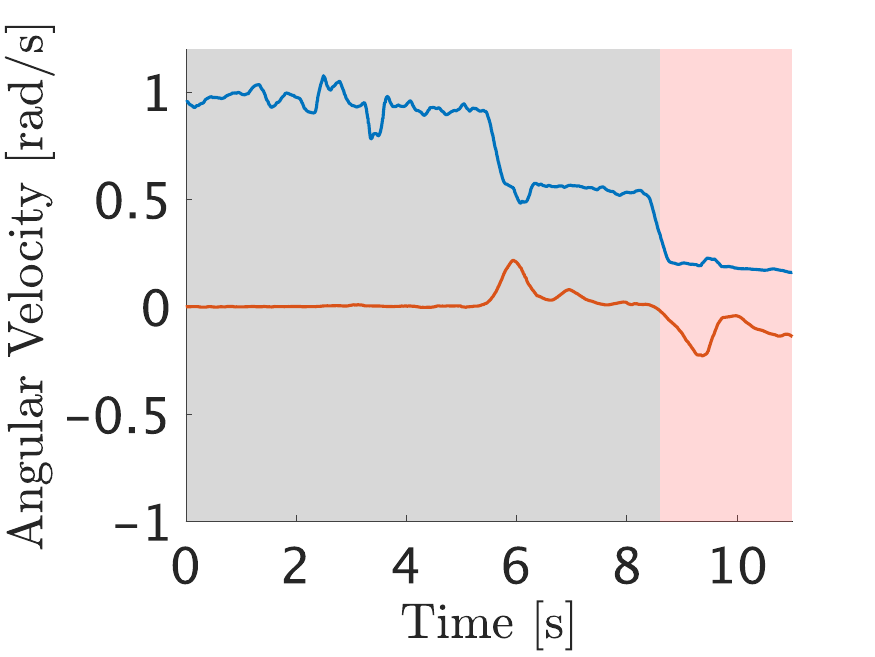}
         \subcaption{Detumbling and caging.}
         \label{fig:exp_mul_w}
     \end{minipage}\\
    \end{tabular}
    % \vspace{-2mm}
    \caption{Base and target angular velocity attenuation in the experiment.}
    \label{fig:exp_target_w}
    \vspace{0mm}
\end{figure}

\begin{figure}[!t]
% \vspace{2mm}
    \begin{tabular}{cc}
     \begin{minipage}[t]{0.48\hsize}
         \centering
         \includegraphics[height=0.75\linewidth]{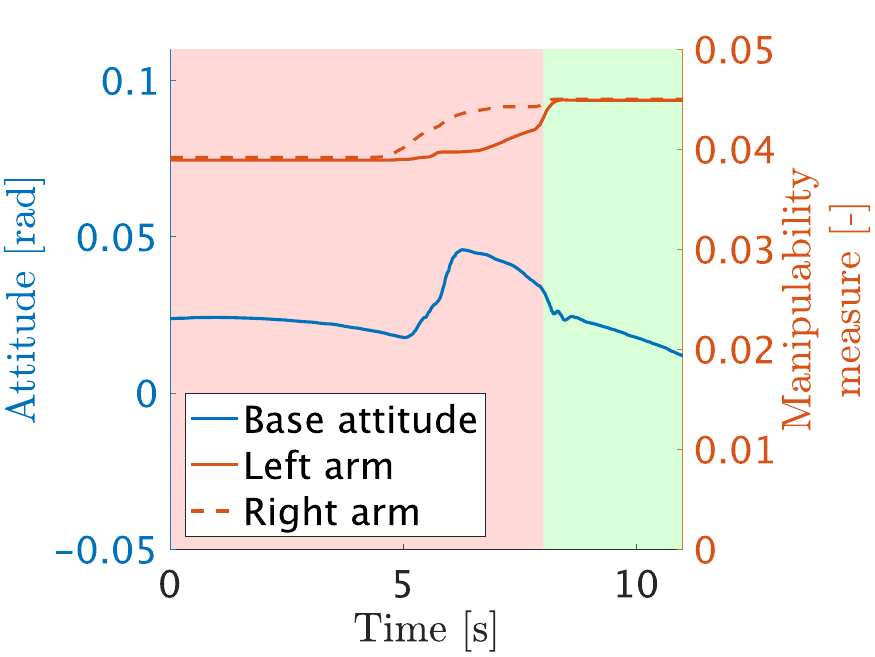}
         \subcaption{Direct caging.}
         \label{fig:exp_dir_stab}
     \end{minipage}&  
      \begin{minipage}[t]{0.48\hsize}
         \centering
         \includegraphics[height=0.75\linewidth]{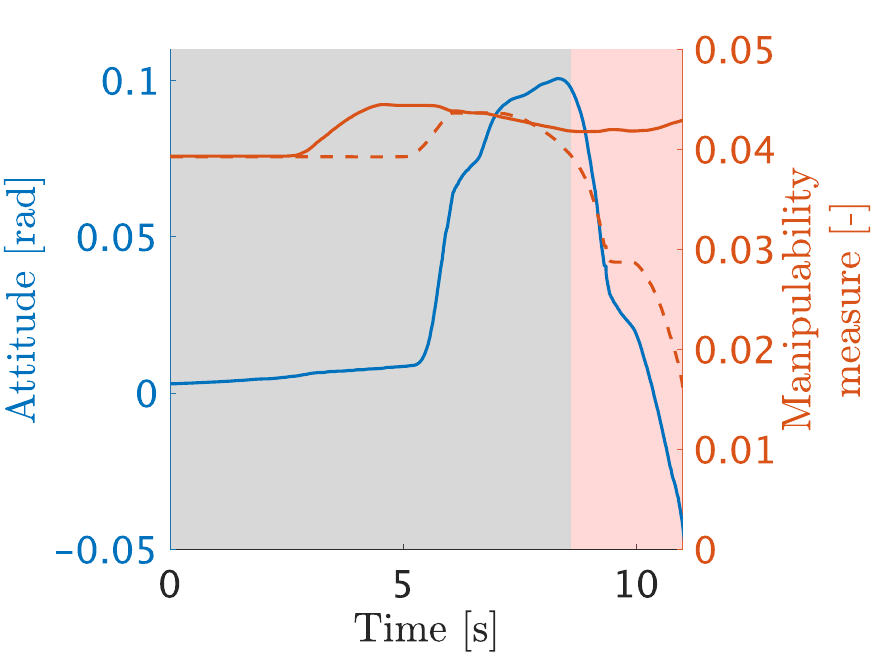}
         \subcaption{Detumbling and caging.}
         \label{fig:exp_mul_stab}
     \end{minipage}\\
    \end{tabular}
    % \vspace{-2mm}
    \caption{Base attitude and measure of manipulability in the experiment.}
    \label{fig:exp_stab}
% \vspace{-3mm}
\end{figure}

As shown in \fig{fig:exp_snapshots}, the robot succeeded in detumbling the target using repeated contacts with impedance control; however, it failed in the following caging phase. This failure occurred because the robot was stacked at a singularity point during the damping phase. \mbox{\fig{fig:exp_stab}}(b) also supports this result by showing the considerably low manipulability of the right arm. This problem can be solved by installing singularity-aware control in the sequence. As it was difficult to model the physics, such as contact dynamics and robot inertia perfectly, the result was different from the simulation wherein the same parameters succeeded in caging. \fig{fig:exp_force} shows that the force applied to the end effector during the detumbling phase was lower compared with that of direct caging. The maximum value was decreased by 89\%, which was a decrease from $4.4$~{[N]} to $0.48$~{[N]}. \fig{fig:exp_force}(a) shows that the force was considerably large in direct caging compared with simulation, which was considered to be because, in direct caging, a small error in position control at high target rotation speeds can induce severe contact, whereas the one-handed detumbling sequence mitigated such occurrences. Furthermore, \fig{fig:exp_target_w} shows that the proposed sequence succeeded in detumbling the target via repeated contact and gradually decreasing the target’s rotational speed, whereas direct caging reduced it rapidly. The final angular velocity of the target was \SI{0.16}{rad/s}, which was below the predefined threshold of \SI{0.5}{rad\ s }, indicating that the detumbling sequence was complete. These results support the simulated results from the experimental aspect that the proposed sequence of detumbling can efficiently distribute the momentum exchange and reduce the impact force between the end effector and debris. 

\section{CONCLUSION}
\label{sec:conclusion}
This study proposed a method combining impedance control and repeated impact-based capture for reliable target detumbling and caging. The authors parametrically analyzed the proposed method in terms of impedance parameters such as virtual mass, damping coefficient, and elastic coefficient, and obtained trends in the successful parameters. This method was validated through both simulations and experiments to successfully detumble debris with a lower force at impact. With our method, the momentum of the target was attenuated through repeated contact, implying that the exchange of momentum between the debris and the satellite was distributed multiple times. The proposed method helped the chaser satellite maintain its attitude using an attitude control unit with a poor response, such as a Reaction Wheel.

In future works, the robustness of the sequence must be verified by testing variations in the weight of the chaser and target. In addition, the speeds necessary for the end effectors of the manipulators must be defined such that the detumbling operation can be conducted depending on the angular velocity of the debris by validating the method in other situations. The desired velocity of the end effector is determined by the contact physics and the allowable impact force with the debris. Further, we aim to conducted a three-dimensional microgravity simulation for more realistic situations and an in-situ estimation of the inertial property of the target based on the reaction force of contacts, which was assumed to be prior knowledge in this study.
% Furthermore, we verified through simulation that a comprehensive sequence where this detumbling technique was adapted to caging successfully captured space debris with lower force impacts. 

%                                   % the textheight of the last page by a suitable amount.
%                                   % This command does not take effect until the next page
%                                   % so it should come on the page before the last. Make
%                                   % sure that you do not shorten the textheight too much.

%%%%%%%%%%%%%%%%%%%%%%%%%%%%%%%%%%%%%%%%%%%%%%%%%%%%%%%%%%%%%%%%%%%%%%%%%%%%%%%%
\section*{Acknowledgment}
The authors thank Koki Abe, Tomoya Matsushita, and Naoki Hase for their invaluable support in the past development of the robot platform and simulation framework.

\bibliography{./IEEEabrv,bibliography.bib}

\end{document}